\documentclass[preprint,5p]{elsarticle}

\usepackage{amsmath,amssymb,amsfonts}
\usepackage{algpseudocode,algorithm}
\usepackage{subcaption}

\journal{Engineering Applications of Artificial Intelligence}

\begin{document}

\sloppy

\begin{frontmatter}

\title{Towards Autonomous Driving of Personal Mobility with Small and Noisy Dataset using Tsallis-statistics-based Behavioral Cloning}

\author[NAIST]{Taisuke Kobayashi\corref{cor}}
\ead{kobayashi@is.naist.jp}
\author[NAIST]{Takahito Enomoto}

\cortext[cor]{Corresponding author}
\address[NAIST]{Division of Information Science, Graduate School of Science and Technology, Nara Institute of Science and Technology, Nara 630-0192, Japan.}

\begin{abstract}

Autonomous driving has made great progress and been introduced in practical use step by step.
On the other hand, the concept of personal mobility is also getting popular, and its autonomous driving specialized for individual drivers is expected for a new step.
However, it is difficult to collect a large driving dataset, which is basically required for the learning of autonomous driving, from the individual driver of the personal mobility.
In addition, when the driver is not familiar with the operation of the personal mobility, the dataset will contain non-optimal data.
This study therefore focuses on an autonomous driving method for the personal mobility with such a small and noisy, so-called personal, dataset.
Specifically, we introduce a new loss function based on Tsallis statistics that weights gradients depending on the original loss function and allows us to exclude noisy data in the optimization phase.
In addition, we improve the visualization technique to verify whether the driver and the controller have the same region of interest.
From the experimental results, we found that the conventional autonomous driving failed to drive properly due to the wrong operations in the personal dataset, and the region of interest was different from that of the driver.
In contrast, the proposed method learned robustly against the errors and successfully drove automatically while paying attention to the similar region to the driver.

\end{abstract}


\end{frontmatter}

\section{Introduction}
\label{sec:introduction}

Recently, autonomous driving of vehicles has reached the stage of practical application.
Autonomous driving is a control problem: e.g. classical control techniques are often employed for keeping lane and distance between vehicles~\cite{suryanarayanan2007appropriate,klanvcar2009wheeled};
planning by model predictive control with vehicle models is also employed for more extensive autonomous driving~\cite{levinson2011towards,williams2018information}.
On the other hand, machine learning is a methodology that can handle the cases where an accurate vehicle model is not available and/or where there is uncertainty in the surrounding environment.
As one of the machine learning technologies, imitation learning, which learns end-to-end mapping from observations to actions (i.e. steering, accelerating, and braking), is mainly utilized with a huge driving dataset~\cite{codevilla2018end,onishi2019end,hawke2020urban}.
In this study, we focus on such an imitation learning technology, which is simpler and more versatile, although its limitations about scalability were reported~\cite{codevilla2019exploring}.

While most of the above-mentioned autonomous driving technologies target general vehicles, the development and widespread use of \textit{personal mobility}, such as electric wheelchairs~\cite{nakajima2017new} and Segways~\cite{nguyen2004segway}, will be accelerated as a next-generation mobility.
The personal mobility is basically intended for short-distance travel and requires the ability to travel in a wide range of situations, not limited to well-developed roads.
In addition, since the personal mobility is developed for personal use, the situations encountered by each driver differ greatly.
That is, it is desirable to tune a controller specialized for each driver rather than acquiring generalized performance by learning from a huge dataset in advance.

The problem considered from this problem setting is the quality of the dataset.
Naturally, the total size of dataset will be small because it is constructed for each driver.
If the driver is not familiar with the operation of the personal mobility, wrong operations will inevitably be included as noises.
Imitation learning on such a small and noisy dataset, called a \textit{personal dataset} in this paper, is known to have significant performance degradation~\cite{argall2009survey,hussein2017imitation}.
For this reason, we have to make imitation learning robust to noise.

Here, we briefly introduce the related work for the noise-robust imitation learning.
A research group of Sugiyama has developed quality-aware imitation learning methods~\cite{wu2019imitation,tangkaratt2020variational}, which estimate the quality of each data to select ones to be optimized.
However, unlike behavioral cloning~\cite{bain1995framework}, which is often used in autonomous driving to learn the direct mapping from observations to actions~\cite{codevilla2018end,onishi2019end,hawke2020urban}, these methods are classified as inverse reinforcement learning~\cite{ng2000algorithms}, which uses reinforcement learning~\cite{sutton2018reinforcement} in combination and requires some trial and error by a non-optimal controller.
R-MaxEnt also estimates the quality of each data through maximum entropy principle~\cite{hussein2021robust}.
Although this method is capable of learning the optimal policy from only a given dataset, the controller is assumed to be for a discrete system, hence, it is not suitable for autonomous driving where the continuous control command is required.
Sasaki and Yamashita have modified the standard behavioral cloning to seek one of the modes of the expert behaviors~\cite{sasaki2021behavioral}.
Although there is no restriction on the controller like above, the controller is desired to be ensemble trained to improve the performance, which increases the computational cost.
Ilboudo et al. have proposed a noise-robust optimizer for the standard behavioral cloning problem~\cite{ilboudo2020robust,ilboudo2021adaptive}.
It checks the gradients used to update the neural networks that approximates the controller, and empirically filters out anomalies, but it is only a safety net and is less effective if there is a lot of noise in the dataset.

Therefore, this paper proposes a simple but yet noise-robust behavioral cloning for the personal mobility.
Specifically, we focus on the fact that the standard behavioral cloning is the minimization problem of the negative log likelihood of the stochastic controller.
By replacing the log likelihood to $q$-log likelihood introduced in Tsallis statistics~\cite{tsallis1988possible,suyari2005law,kobayashi2020q}, the behavioral cloning can easily adjust its noise robustness in accordance with a real parameter, $q$.
This replacement can be interpreted as a nonlinear transformation of the log likelihood, and the gradient naturally disappears in noisy data where the log likelihood becomes small.
As a result, each data is implicitly weighted to imitate only high quality data, resulting in obtaining the noise robustness.

In order to validate the proposed method, we employ a visualization technique for the inputs (more specifically, the region of interest in the inputted image) that is strongly involved in the controller, so-called VisualBackProp~\cite{bojarski2018visualbackprop}.
This allows us to qualitatively assess whether the driver and the learned controller have a common region of interest.
However, we empirically found that the original VisualBackProp sometimes fails to extract the region of interest appropriately due to noise, which causes extreme features values.
In addition, although conventional techniques are for convolutional neural networks (CNNs)~\cite{krizhevsky2012imagenet,lecun2015deep}, in many cases, the features are further shaped by multiple fully connected networks (FCNs) after the CNNs.
These FCNs are ignored in the original VisualBackProp, thus ignoring the features that contribute more directly to the controller.
Therefore, as an additional minor contribution, we modify the implementation of VisualBackProp to improve these shortcomings.

Experiments using an electric wheelchair as one of the personal mobilities are conducted for the verification of the proposed method.
The personal dataset contains driving corners, stopping in front of the stop sign, and zigzagging and/or non-stopping as noise.
Although the standard behavioral cloning fails to imitate the stopping operation due to the adverse effects by the noisy data, the proposed method successfully imitates all the operations by excluding the noisy data.
In addition, the modified VisualBackProp is able to properly extract the stop sign (and objects in a shelf to guide driving a corner) as the region of interest, which is naturally similar to that of the driver.
As a consequence, the proposed method achieves the autonomous driving of the personal mobility even with the small and noisy personal dataset, while extracting the driver-like region of interest.

\section{Conventional methods and their problems}
\label{sec:conventional}

\subsection{Behavioral cloning}

\begin{figure}[tb]
    \centering
    \includegraphics[keepaspectratio=true,width=0.99\linewidth]{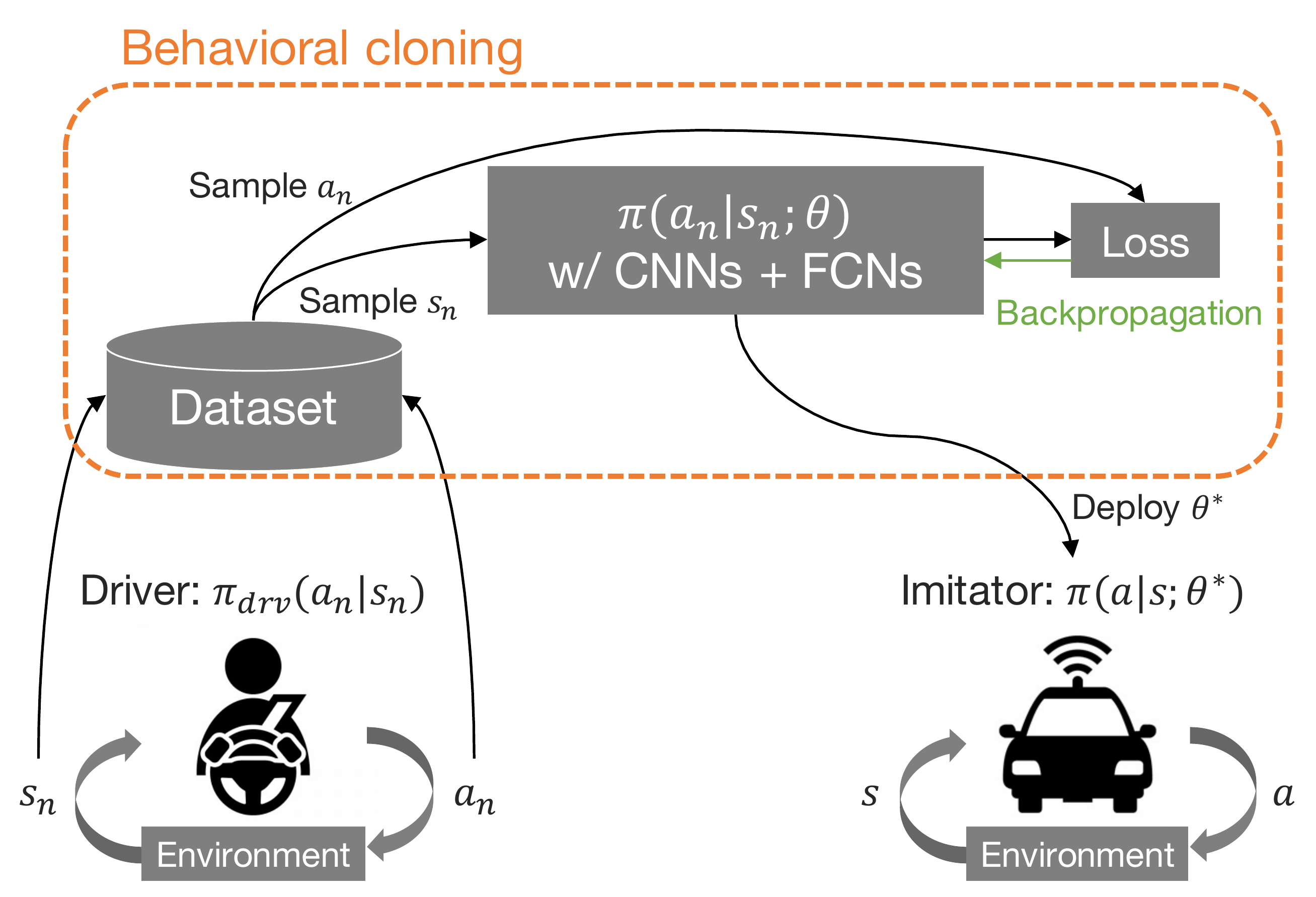}
    \caption{Overview of behavioral cloning:
    at first, the dataset with the pairs of $(s_n, a_n)$ is collected by an expert (often a human);
    the parameters set $\theta$ are optimized using the dataset in an offline manner;
    after optimization, $\theta^*$ is deployed to imitate the expert operations in the real environment.
    }
    \label{fig:arch_bc}
\end{figure}

Behavioral cloning is one of the most popular imitation learning methods~\cite{bain1995framework}.
Under Markov process, a dataset including $N$ pairs of expert operations $a$ over observed states $s$, $D = \{(s_n, a_n)\}^N_{n=1}$, is built.
We consider learning a stochastic controller $\pi(a \mid s; \theta)$ with $\theta$ the parameters set, assuming that $D$ includes stochastic operations, especially when the expert is human(s).
Since $\pi$ is a distribution model parameterized by $\theta$ (e.g. normal distribution), the following minimization of the negative log likelihood with $D$ is employed for optimization of $\theta$.
\begin{align}
    \theta^* &= \arg\min_{\theta} \mathcal{L}(\theta)
    \nonumber \\
    \mathcal{L}(\theta) &= - \mathbb{E}_{(s_n, a_n) \sim D}[\ln \pi(a_n \mid s_n)]
    \label{eq:loss_vanilla}
\end{align}
This optimization problem is basically solved by stochastic gradient descent (e.g. Adam~\cite{kingma2014adam}) when $\pi$ is approximated by neural networks, i.e. $\theta$ contains network weights and biases.
The above process is illustrated in Fig.~\ref{fig:arch_bc}.

$\pi^* = \pi(a \mid s; \theta^*)$ obtained through this problem is optimized to equally represent all the data in $D$ as well as possible.
If $D$ is ideal and huge, $\pi^*$ after deployed should achieve the expert imitation properly.
However, if $D$ contains incorrect operations, as this paper deals with, a risk of interference, such as requiring different $a$ for the same $s$, will be increased, leading to failure of the proper expert imitation.
In addition, the smaller $D$ is, the more the effect of such noise becomes apparent.

\subsection{VisualBackProp}

\begin{figure}[tb]
    \centering
    \includegraphics[keepaspectratio=true,width=0.99\linewidth]{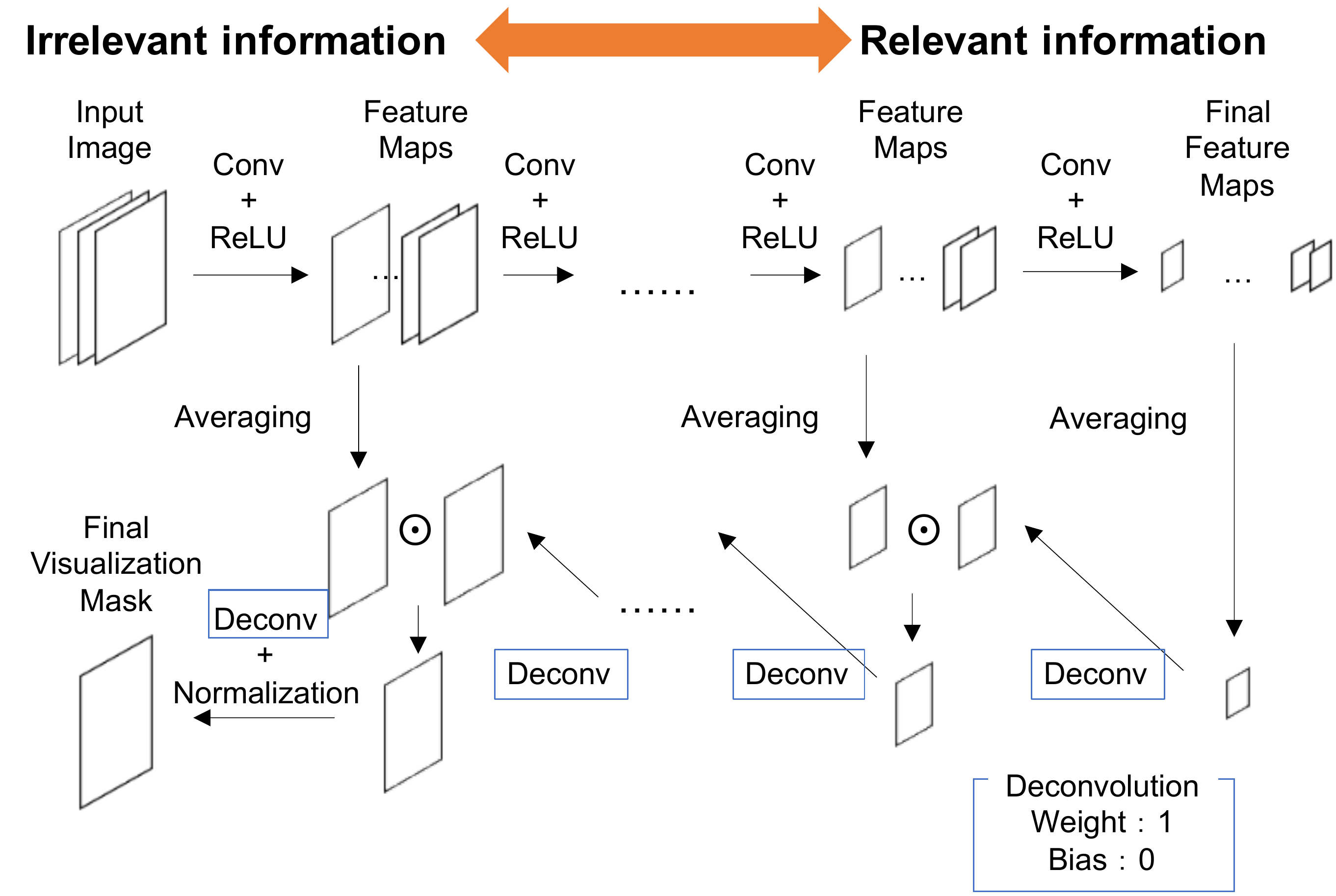}
    \caption{Flow of VisualBackProp:
    by backpropagating the features at the respective CNN layers, the attention map (or mask image) is extracted to reveal the region of interest for the output.
    }
    \label{fig:arch_vbp}
\end{figure}

VisualBackProp is one of the methods to enhance the interpretability of the outputs obtained through CNNs~\cite{bojarski2018visualbackprop}.
Following the process below (illustrated in Fig.~\ref{fig:arch_vbp}), an attention map (or a mask image) is generated from features obtained in the respective CNNs corresponding to the input image.
This attention map identifies the region of interest that contributes significantly to the output.
\begin{enumerate}
    \item Get the feature averaged in the channel direction, $x_L$, from the $L$-th CNN layer closest to the output layer, and set it as $m_L$.
    \item Pass $m_L$ through the deconvolution layer~\cite{zeiler2011adaptive} with weights of one and a bias of zero, as $\tilde{x}_L$, to match the feature size of the $L-1$-th layer.
    \item Compute the element product of $\tilde{x}_L$ and $x_{L-1}$ as $m_{L-1}$.
    \item Decrement $L$ and repeat the steps 2 and 3 until reaching the first CNN layer.
    \item Normalize $m_1$ to make all the components within $[0, 1]$.
\end{enumerate}

When ReLU functions are employed as the activation functions for the respective CNNs, their features become non-negative, and it can be expected that the mask image, in which only pixels with high contribution have non-zero values only by the element product, is extracted.
However, if some of the features are with excessive values, their effects will remain unless the counterpart of the element product is perfectly zero, and they may overwrite other features.
Excessive values of some features are prone to occur when noise is mixed in with the input, and therefore, we have to consider this problem in this paper.
In addition, the feature obtained by passing through CNNs is not directly converted to the output, but may be further formatted by FCNs.
Since VisualBackProp ignores the effects of FCNs, it is difficult to say that it truly generates the region of interest that contributes to the output.

\section{Noise-robust behavioral cloning}
\label{sec:proposal1}

\begin{figure*}[tb]
    \begin{subfigure}[b]{0.33\linewidth}
        \centering
        \includegraphics[keepaspectratio=true,width=\linewidth]{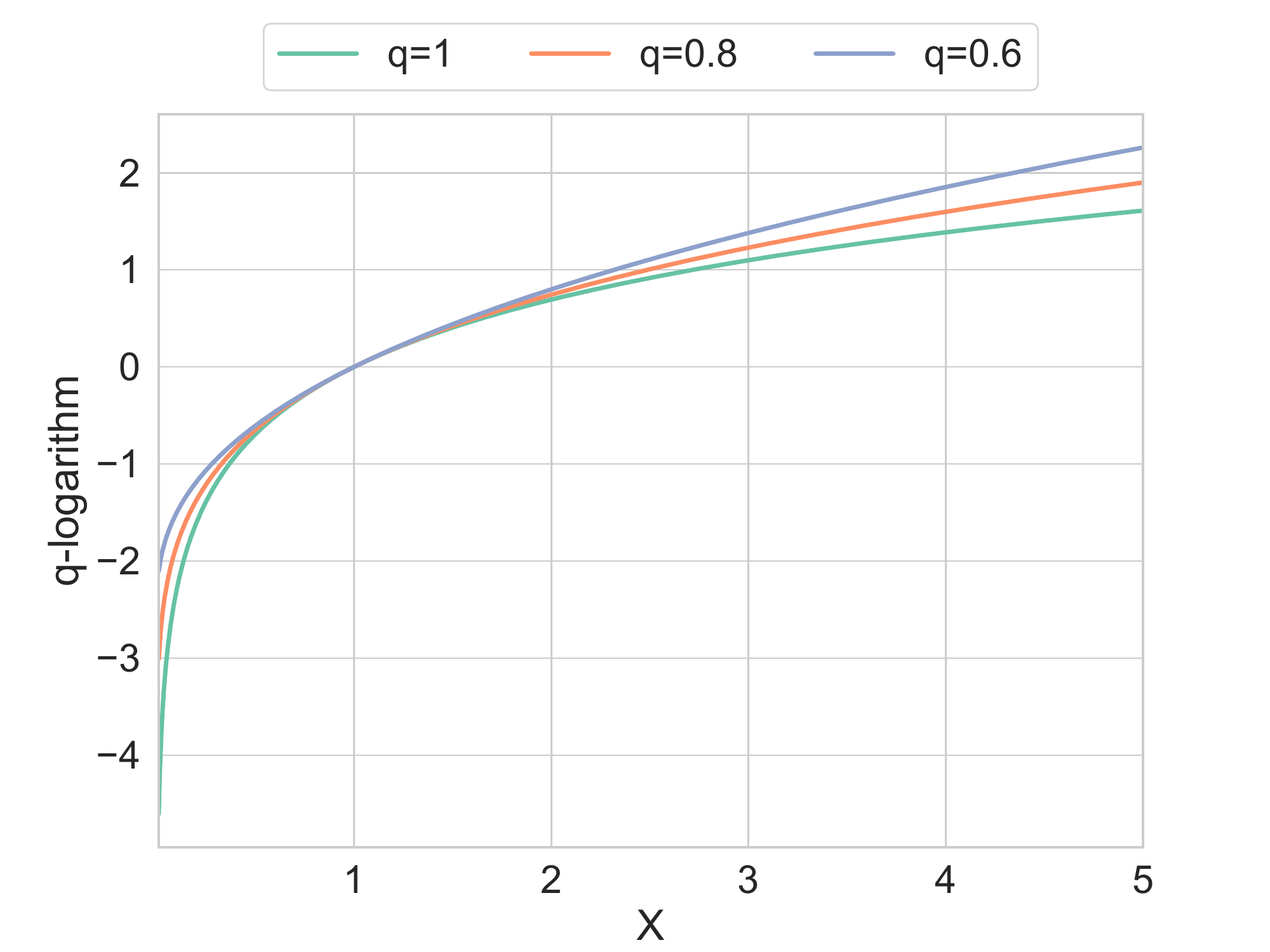}
        \subcaption{$x$ vs. $\ln_q(x)$}
        \label{fig:anl_nrbc_lnqx}
    \end{subfigure}
    \begin{subfigure}[b]{0.33\linewidth}
        \centering
        \includegraphics[keepaspectratio=true,width=\linewidth]{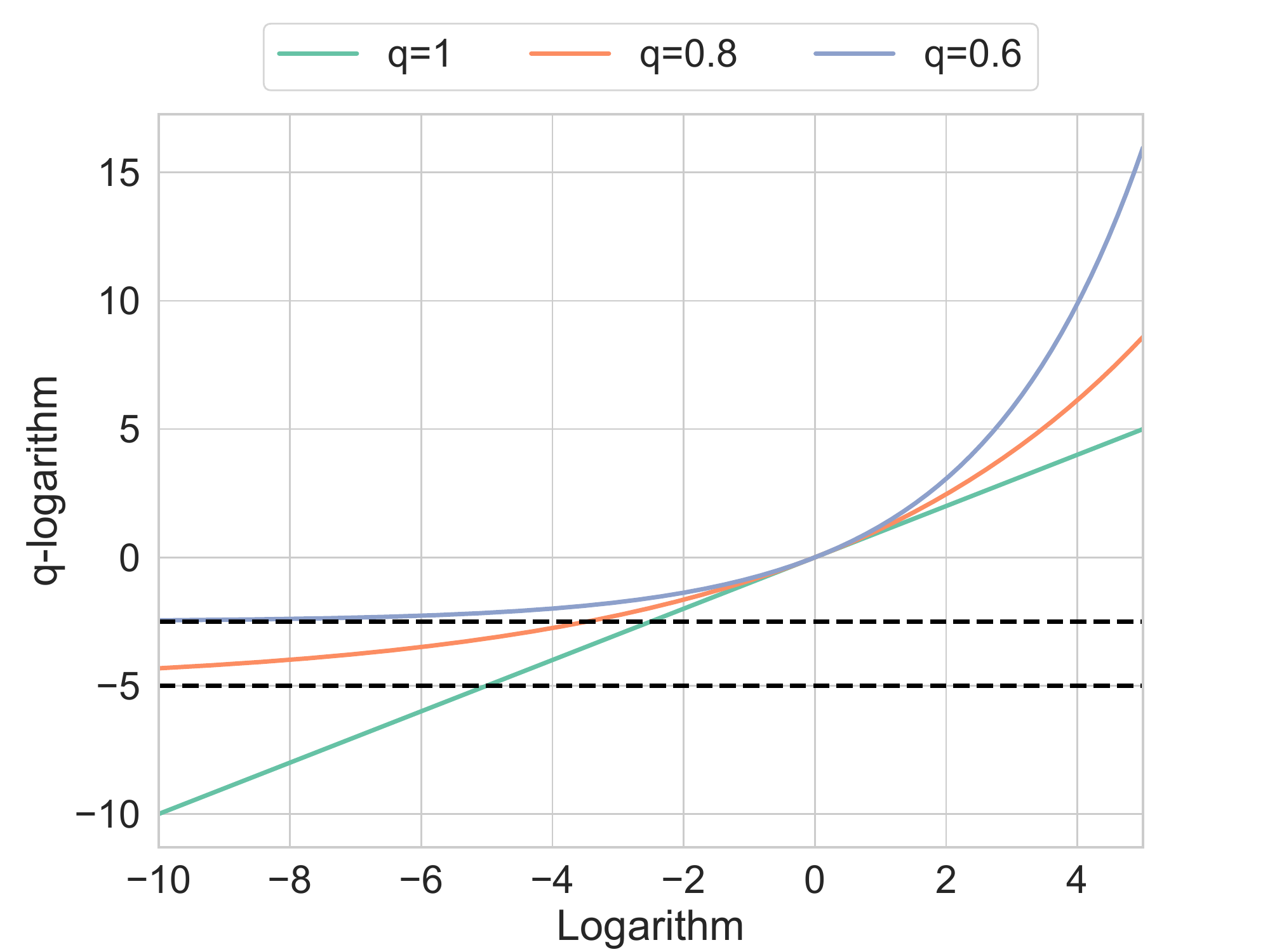}
        \subcaption{$\ln(x)$ vs. $\ln_q(\ln(x))$}
        \label{fig:anl_nrbc_lnqln}
    \end{subfigure}
    \begin{subfigure}[b]{0.33\linewidth}
        \centering
        \includegraphics[keepaspectratio=true,width=\linewidth]{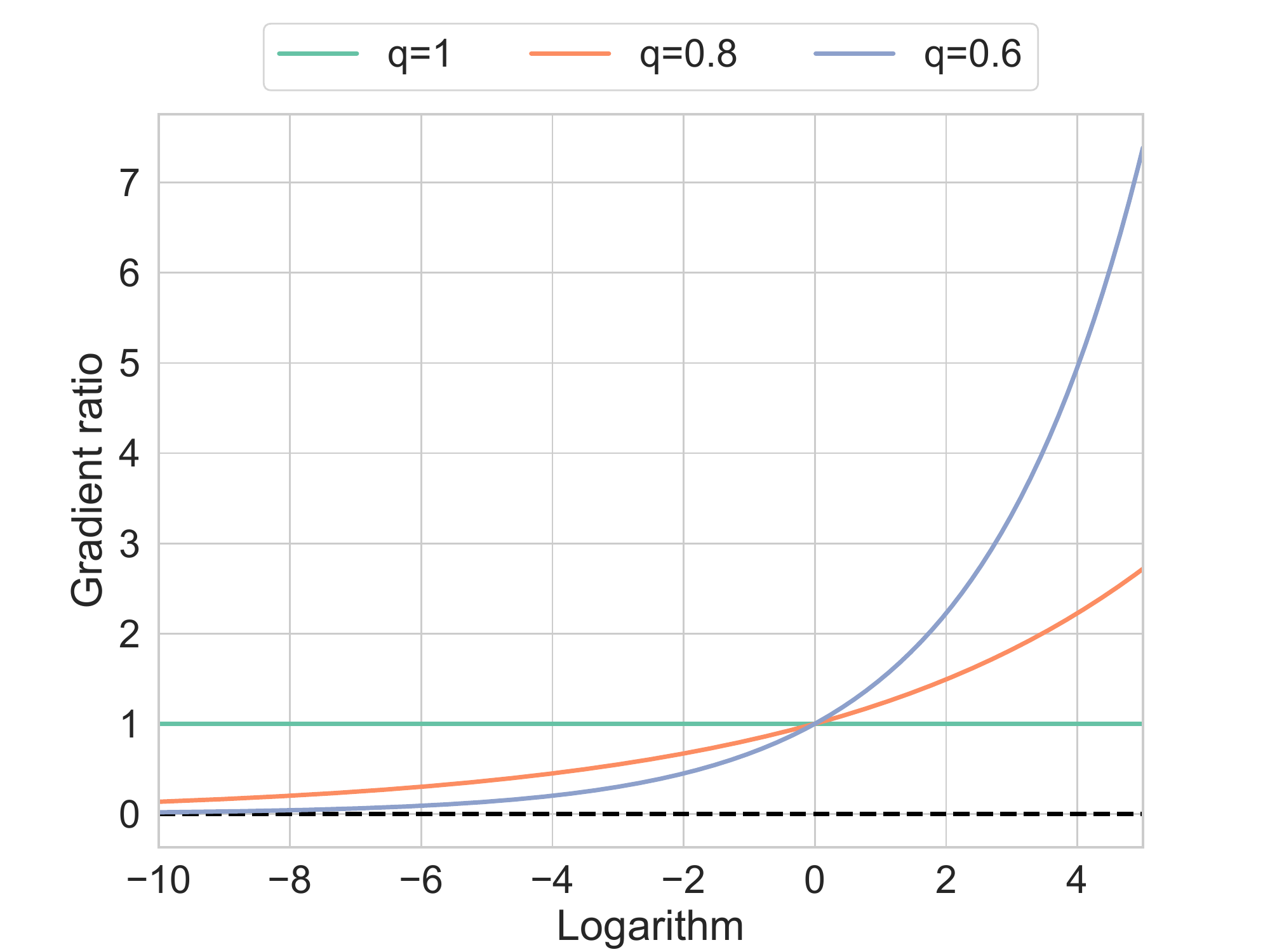}
        \subcaption{$\ln(x)$ vs. $\rho(\ln(x))$}
        \label{fig:anl_nrbc_ratio}
    \end{subfigure}
    \caption{Analyses of noise robustness in the proposed method:
    (a) since the monotonic increasing function $\ln_q(x)$ with $q < 1$ is larger than or equal to $\ln(x)$ (see eq.~\eqref{eq:q_ineq1}), the gradient of $\ln_q(x)$ at $x < 1$ is smaller than that of $\ln(x)$ (see eq.~\eqref{eq:q_ineq2});
    (b) $\ln_q(x)$ can be defined as the function of $\ln(x)$ (see eq.~\eqref{eq:q_log2}), and its gradient with $q < 1$ converges to zero at $\ln(x) \to - \infty$;
    (c) since the ratio of the gradients of $\ln_q(x)$ and $\ln(x)$, $\rho(\ln(x))$, is expressed as an exponential function (see eq.~\eqref{eq:ratio}), data can be exponentially weighted according to their losses and the noisy data with larger losses would be ignored.
    }
    \label{fig:anl_nrbc}
\end{figure*}

\subsection{Tsallis statistics}

Tsallis statistics refers to the organization of mathematical functions and associated probability distributions proposed by Tsallis~\cite{tsallis1988possible,suyari2005law}.
This concept is organized based on $q$-deformed exponential and logarithmic functions, which are extensions of general exponential and logarithmic functions by real number $q \in \mathbb{R}$.
Tsallis statistics has various properties, and machine learning methods that take advantage of these properties have been proposed, such as~\cite{kobayashi2020q}.
We introduce the following $q$-logarithm for our method.

The $q$-logarithm, $\ln_q(x)$ with $x > 0$, is given as follows:
\begin{align}
    \ln_q(x) =
    \begin{cases}
        \ln(x) & q = 1
        \\
        \cfrac{x^{1 - q} - 1}{1 - q} & q \neq 1
        \\
    \end{cases}
    \label{eq:q_log}
\end{align}
where $q$ gives its shape.
Regardless of $q$, $\ln_q(x)$ belongs to monotonic increasing function.

\subsection{Formulation with $q$-log likelihood}

The proposed noise-robust behavioral cloning can easily be derived with eqs.~\eqref{eq:loss_vanilla} and~\eqref{eq:q_log}.
Specifically, given $q \leq 1$, the log likelihood in eq.~\eqref{eq:loss_vanilla} is replaced by the $q$-log likelihood as follows:
\begin{align}
    \theta_q^* &= \arg\min_{\theta} \mathcal{L}_q(\theta)
    \nonumber \\
    \mathcal{L}_q(\theta) &= - \mathbb{E}_{(s_n, a_n) \sim D}[\ln_q \pi(a_n \mid s_n)]
    \label{eq:loss_robust}
\end{align}
When $q=1$, this can be reverted to the standard behavioral cloning.
Note that since $\ln_q(x)$ including $\ln(x)$ is the monotonic increasing function as mentioned before, the direction of learning itself is invariant with this replacement.

\subsection{Analysis of noise robustness}

We show from two analyses why this formulation is robust to noise.
We notice again that when $\pi$ is represented by neural networks, the behavioral cloning problem is solved by stochastic gradient descent (e.g. Adam~\cite{kingma2014adam}), hence, the gradient property is important for the analyses.
The following analyses can be illustrated in Fig.~\ref{fig:anl_nrbc}.

Before the analyses, we assume that the number of noisy data is few compared to the number of normal data.
In addition, the loss (i.e. the negative ($q$-)log likelihood) for the noisy data is larger than the others.
This is a natural assumption since the limited resources ($\theta$) are allocated to represent the normal and majority data and the remaining has no enough capability to do the noisy data, although it inhibits learning the normal data.

First, for $\ln_q(x)$, the following inequality is satisfied.
\begin{align}
    \ln_{q_1}(x) \leq \ln_{q_2}(x)
    \ \mathrm{s.t. }\ q_1 > q_2
    \label{eq:q_ineq1}
\end{align}
The equality is valid only when $x=1$.
The special case of this inequality is $\ln(x) \leq \ln_q(x)$ with $q < 1$.
In order to satisfy this inequality while matching on $x=1$, the following two inequalities must be satisfied since $\ln_q(x)$ is monotonic.
\begin{align}
    \begin{cases}
        \cfrac{d \ln(x)}{dx} > \cfrac{d \ln_q(x)}{dx} & x < 1
        \\
        \cfrac{d \ln(x)}{dx} < \cfrac{d \ln_q(x)}{dx} & x > 1
    \end{cases}
    \label{eq:q_ineq2}
\end{align}
Although $x < 1$ is often the case for the noisy data with a large loss, the gradient of the proposed method becomes small.
That is, it does not try to reduce its loss relative to the other normal data with $x > 1$.
The proposed method therefore achieves learning with priority on the normal data.

For a more precise analysis, we derive the ratio of the gradients for $\ln(x)$ and $\ln_q(x)$ as $\rho(\ln(x))$.
This can be easily gained by representing $\ln_q(x)$ as a function of $\ln(x)$.
\begin{align}
    \ln_q(\ln(x)) =
    \begin{cases}
        \ln(x) & q = 1
        \\
        \cfrac{\exp\{(1 - q)\ln(x)\} - 1}{1 - q} & q \neq 1
        \\
    \end{cases}
    \label{eq:q_log2}
\end{align}
where $x = \exp(\ln(x))$ is utilized.
Its gradient for $\ln(x)$ (i.e. the gradient ratio or weight $\rho(\ln(x))$) can be analytically given as follows:
\begin{align}
    \rho(\ln(x)) = \cfrac{d \ln_q(\ln(x))}{d \ln(x)} = \exp\{(1 - q)\ln(x)\}
    \label{eq:ratio}
\end{align}
Note that this equation can cover the case with $q=1$.
This means that each data is exponentially weighted according to its own loss.
That is, with $x < 1$ for the noisy data, $\rho(\ln(x))$ converges to zero, hence the noisy data would be ignored.
In addition, the smaller $q$ yields the faster the convergence of $\rho(\ln(x)) \to 0$.
However, please note that if $q$ is too small, even the normal data will be ignored, and therefore, we have to tune $q$ appropriately by checking test data.

\section{Modified VisualBackProp}
\label{sec:proposal2}

\begin{figure*}[tb]
    \centering
    \includegraphics[keepaspectratio=true,width=0.99\linewidth]{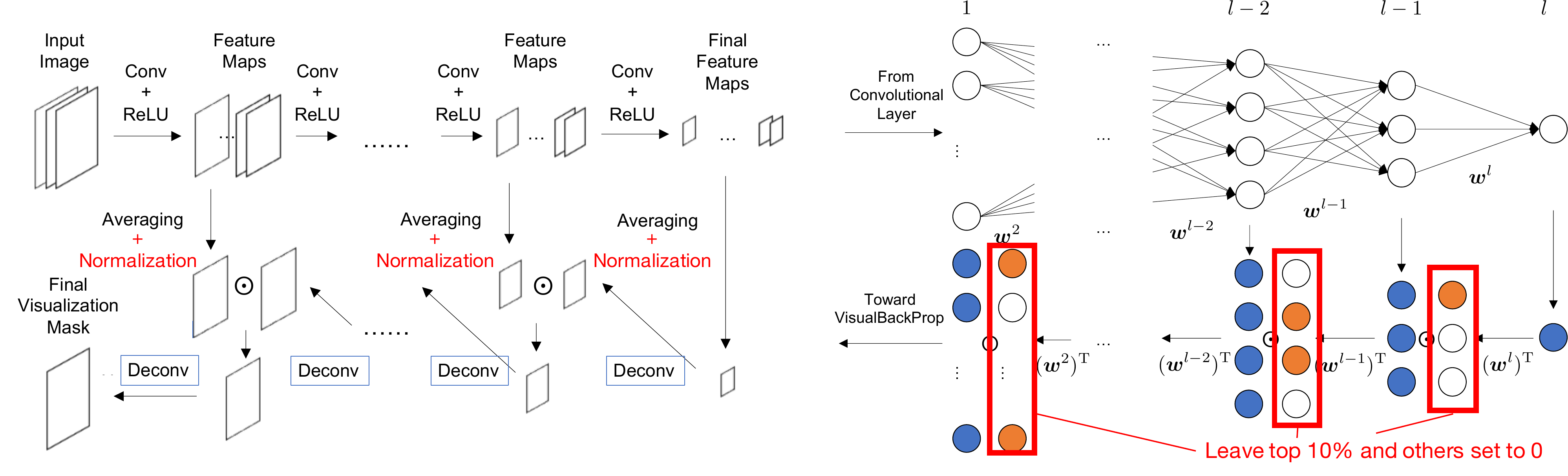}
    \caption{Modified version of VisualBackProp:
    to alleviate the effects of large features, all the features are normalized before the element product;
    to visualize the effects of FCNs, the features in FCNs are also backpropagated through the sparse connection matrices.
    }
    \label{fig:arch_vbp_mod}
\end{figure*}

\subsection{Normalization of intermediate features}

This section implements a minor fix to VisualBackProp~\cite{bojarski2018visualbackprop}.
The modified VisualBackProp is shown in Fig.~\ref{fig:arch_vbp_mod}.

In the original VisualBackProp, the features with large values are back-propagated to the input mask, and the resulting region of interest may look like a blurred image of the entire input image and cannot be limited to a specific region.
Although it is a naive approach, we can alleviate this problem by normalizing all components of each feature $x_L$ to be $[0, 1]$.
This process is expected to make all components of the mask $m_L$ in each layer also $[0, 1]$, resulting in that unnecessary information will be removed as zero and important information will remain as one.

\subsection{Backpropagation from fully connected networks}

As another issue in the original VisualBackProp, we consider the effects of FCNs after CNNs.
The main difference between FCNs and CNNs is the deconvolution process, except that the features of FCNs can be backpropagated by the same procedure as for CNNs.

The deconvolution process for FCNs can be represented by transposing the weight matrix.
However, if all the weights are set to 1 as in the case of CNNs, all the components of the expanded feature will be the sum of the feature components before the expansion because they are all combined, unlike CNNs.
To avoid this problem, we introduce a sparse connection matrix where only the top 10\% of the forward weight matrix is 1 and the rest is 0.
This allows us to backpropagate the FCN features with high importance (i.e. with large weights).

\section{Experiment}
\label{sec:experiment}

\begin{figure}[tb]
    \centering
    \includegraphics[keepaspectratio=true,width=0.99\linewidth]{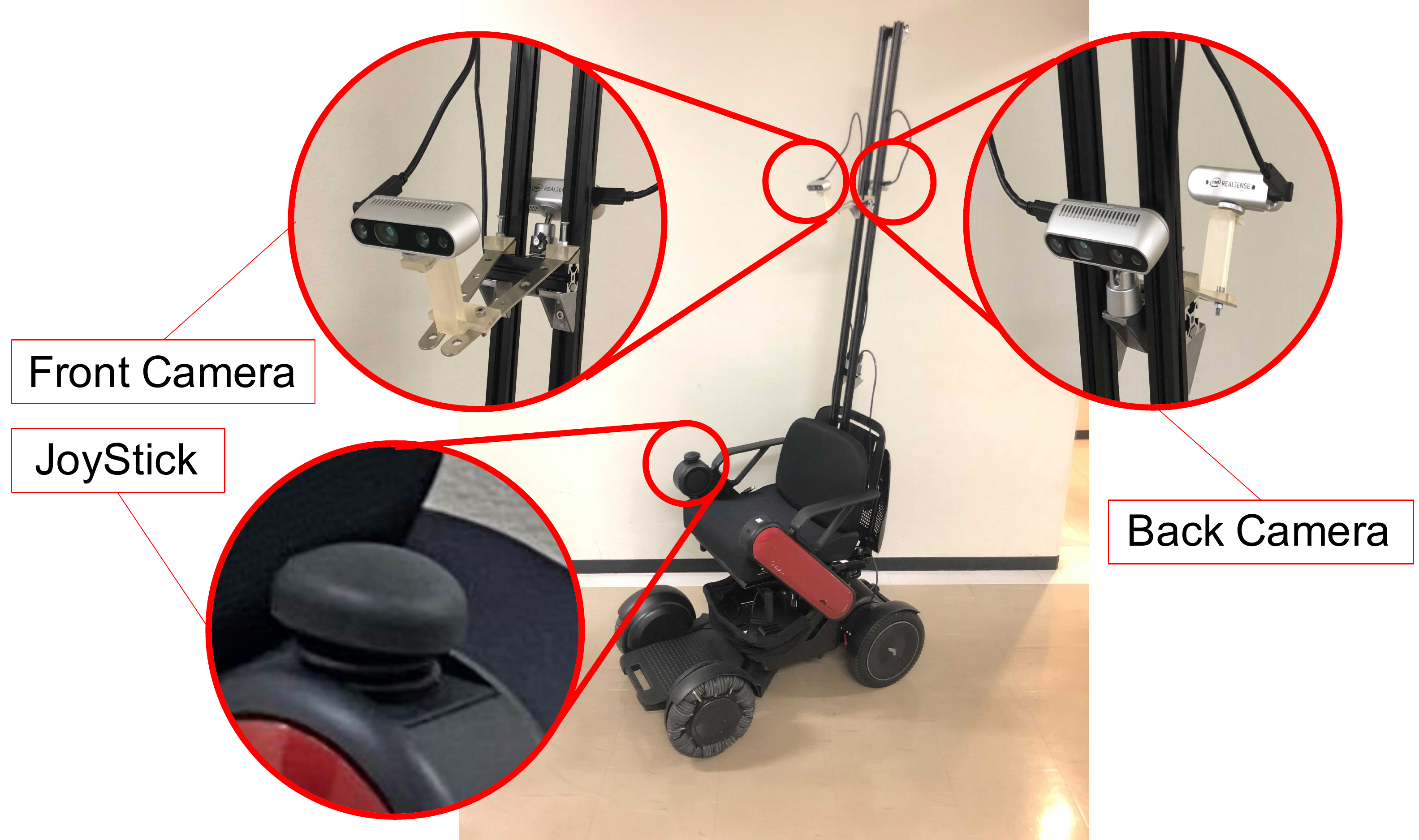}
    \caption{Electric wheelchair as one of the personal mobility:
    the joystick commands can be sent from ROS2, so that the same control system can be shared between the driver and the system;
    the front camera is positioned and angled so that it overlaps with the driver field of view as much as possible.
    }
    \label{fig:exp_wheel}
\end{figure}

\begin{figure}[tb]
    \centering
    \includegraphics[keepaspectratio=true,width=0.99\linewidth]{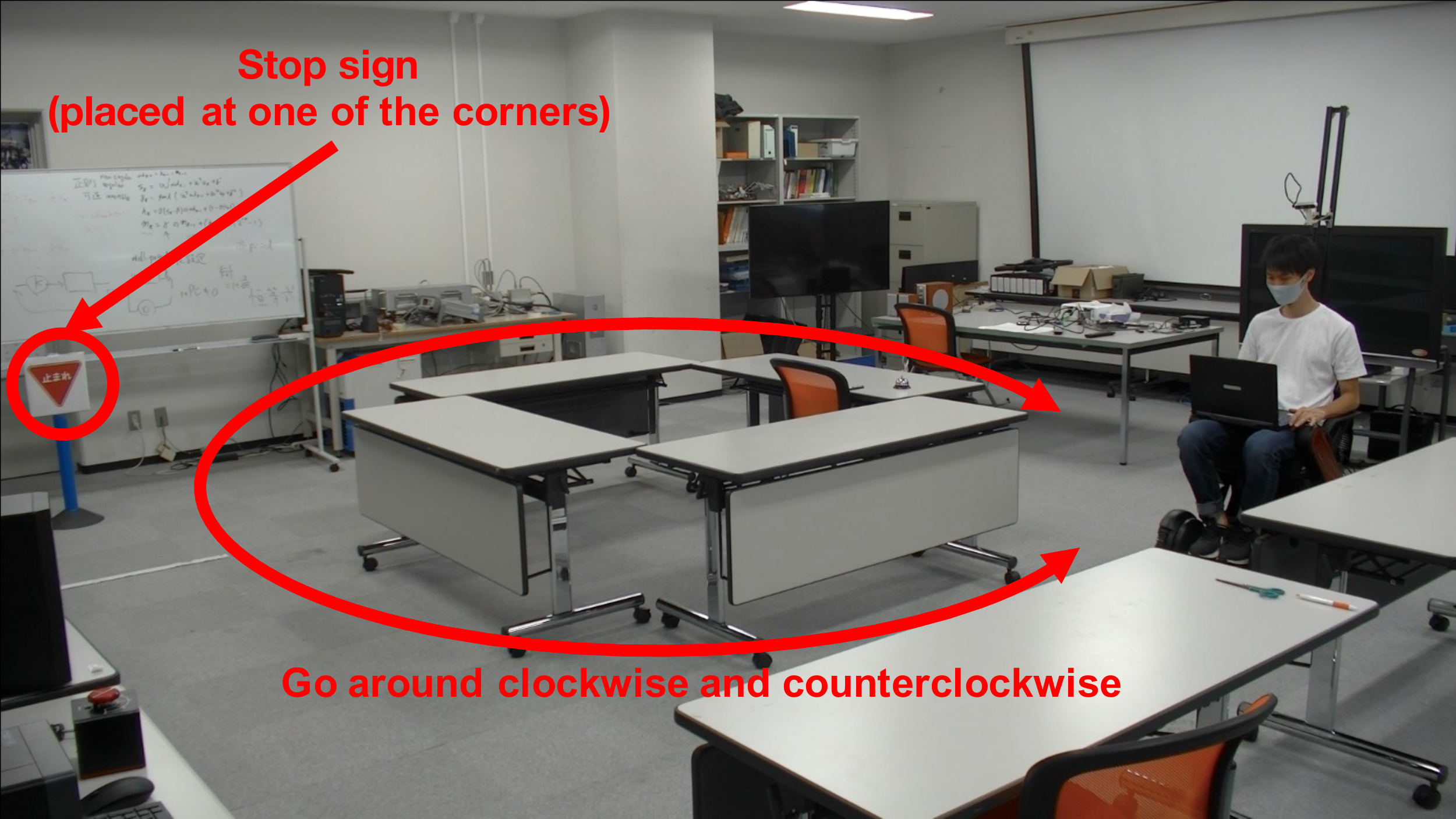}
    \caption{Traveled course:
    the driver goes around a rectangular course with everyday objects placed in a messy manner around to make it easy to extract the visual features;
    a stop sign is placed diagonally opposite the starting point, and the wheelchair needs to stop before it.
    }
    \label{fig:exp_course}
\end{figure}

\begin{figure*}[tb]
    \centering
    \includegraphics[keepaspectratio=true,width=0.99\linewidth]{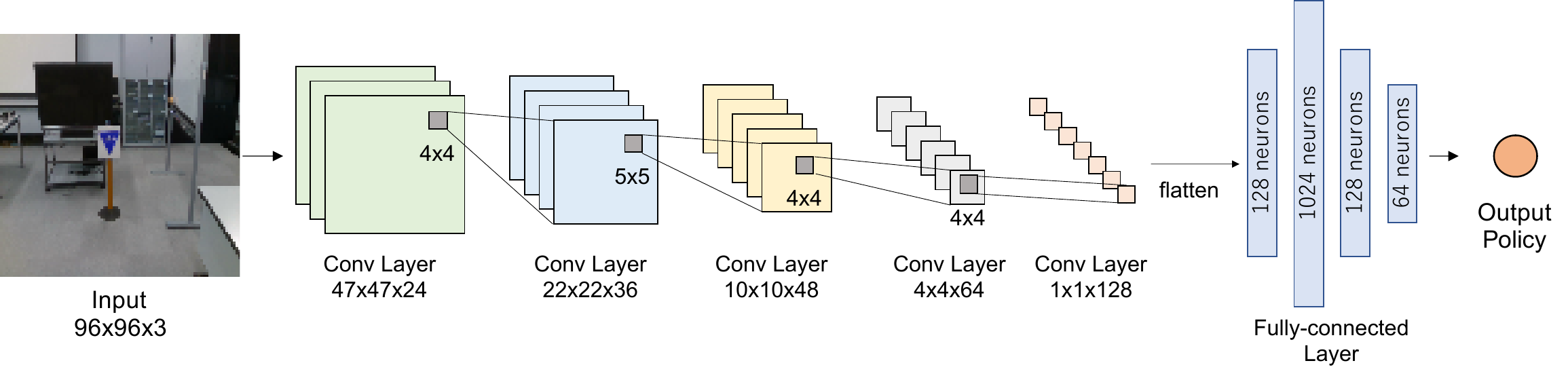}
    \caption{Network architecture to output the controller (policy):
    a compressed image is fed into this and passes through five CNNs until the height and width are one;
    the 128 channel features are considered as firings of the neurons and fed into three FCNs;
    finally the parameters (mean, variance) of a two-dimensional diagonal normal distribution are outputted.
    }
    \label{fig:arch_network}
\end{figure*}

\begin{figure}[tb]
    \centering
    \includegraphics[keepaspectratio=true,width=0.99\linewidth]{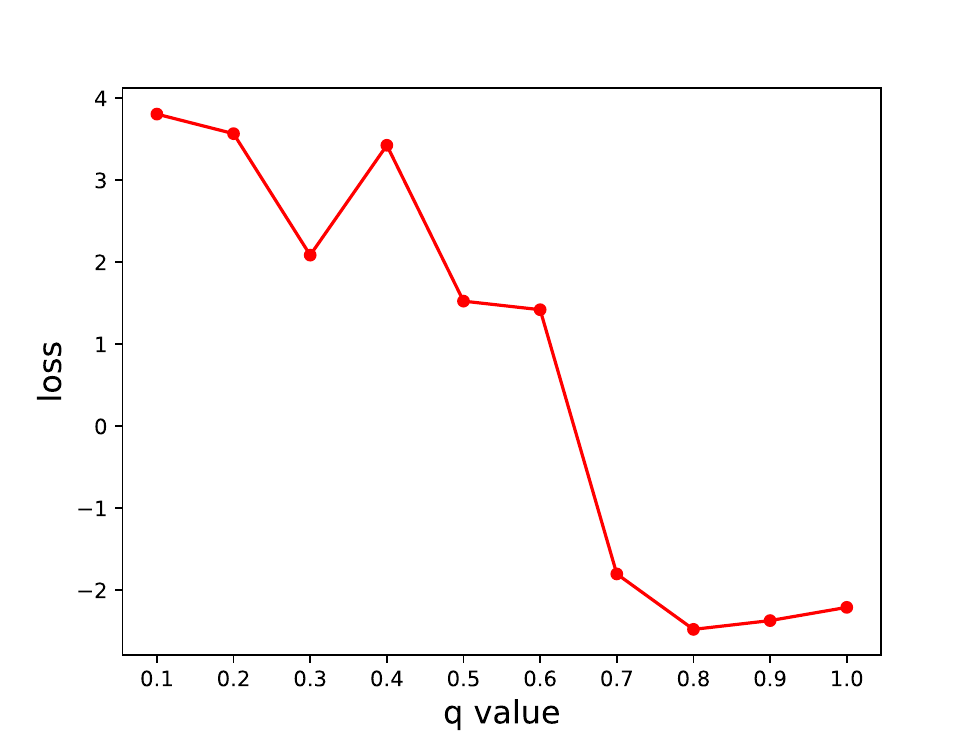}
    \caption{Evaluation of the effect of $q$:
    when $q$ is too small, the proposed method collapses the mapping from state to action;
    with moderate $q$ ($q=0.8$ in this case), the proposed method outperforms the standard behavioral cloning.
    }
    \label{fig:result_qvals}
\end{figure}

\begin{figure}[tb]
    \centering
    \includegraphics[keepaspectratio=true,width=0.99\linewidth]{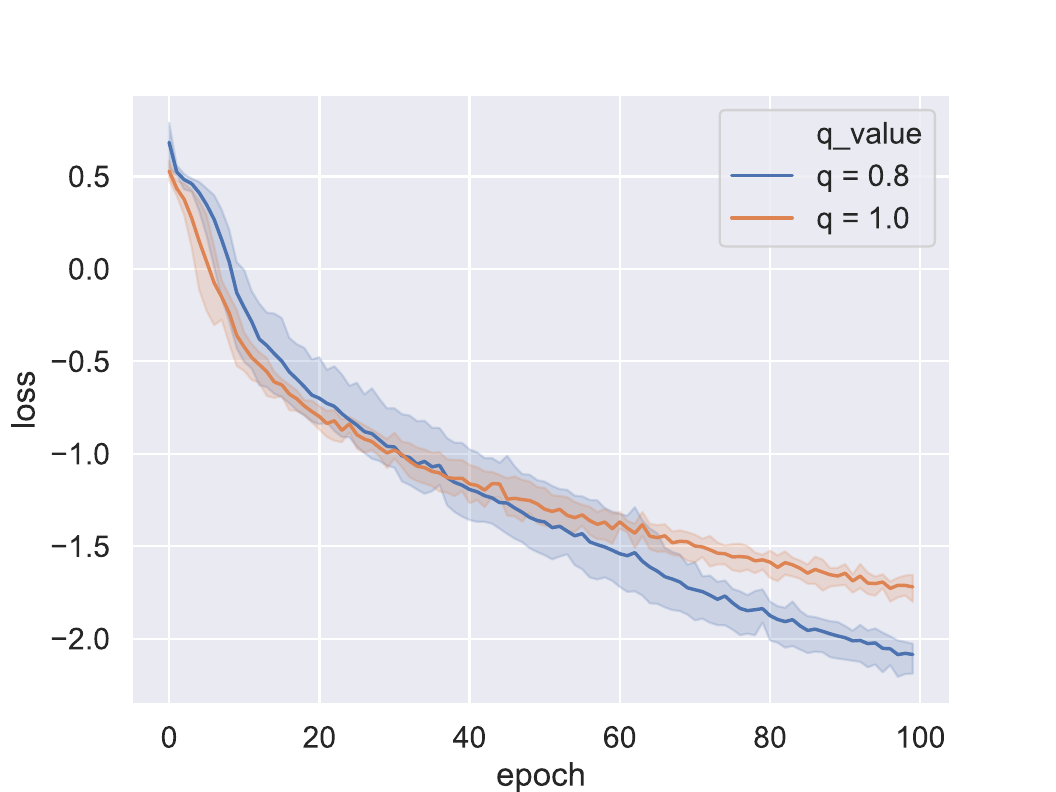}
    \caption{Learning curves for $q=0.8$ as the proposed method and $q=1$ as the conventional method:
    the update speed of the proposed method is rather slow in the early stage, probably because the overall loss is large and many data are with small weights (i.e. gradient ratios $\rho$);
    thanks to the noise robustness, the proposed method outperforms the conventional method from the middle stage and succeeds in reducing the loss steadily thereafter.
    }
    \label{fig:result_loss}
\end{figure}

\begin{figure*}[tb]
    \begin{subfigure}[b]{0.99\linewidth}
        \centering
        \includegraphics[keepaspectratio=true,width=\linewidth]{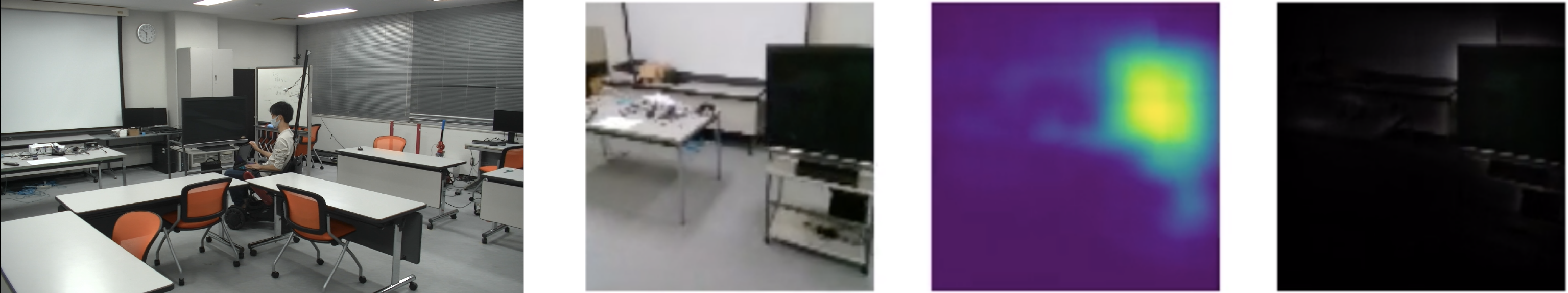}
        \subcaption{At corner}
        \label{fig:snap_vanilla_1}
    \end{subfigure}
    \begin{subfigure}[b]{0.99\linewidth}
        \centering
        \includegraphics[keepaspectratio=true,width=\linewidth]{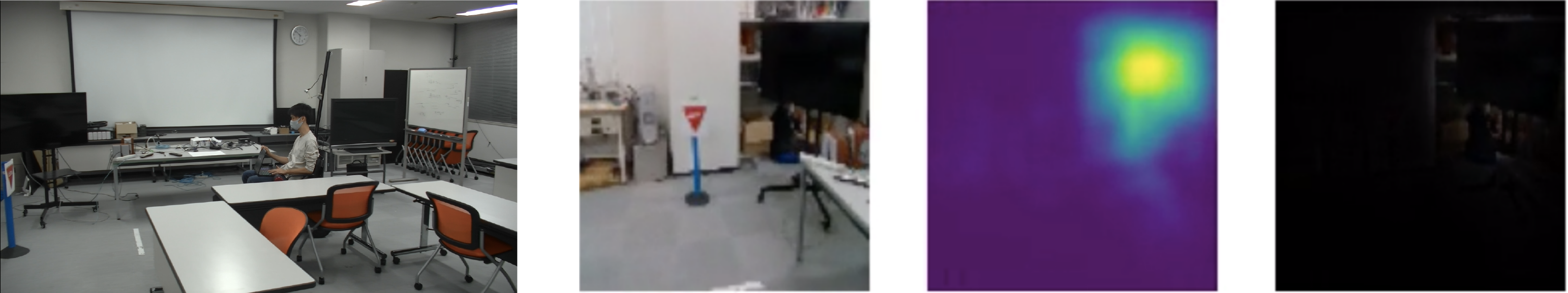}
        \subcaption{In front of stop sign}
        \label{fig:snap_vanilla_2}
    \end{subfigure}
    \begin{subfigure}[b]{0.99\linewidth}
        \centering
        \includegraphics[keepaspectratio=true,width=\linewidth]{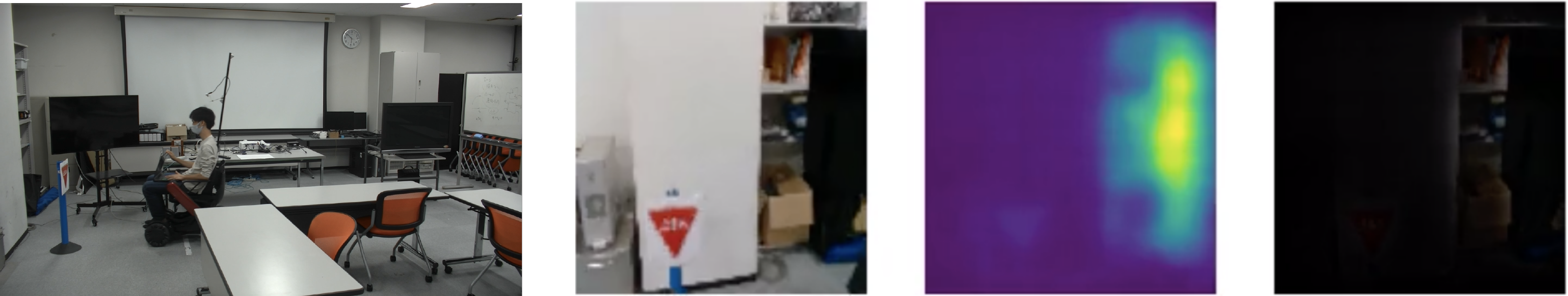}
        \subcaption{When stopped}
        \label{fig:snap_vanilla_3}
    \end{subfigure}
    \caption{Snapshots and the corresponding regions of interest when demonstrated by the conventional method:
    the wheelchair succeeded in turning the left at the corner noisily, but it completely missed the stop sign and eventually crossed the stop line.
    }
    \label{fig:snap_vanilla}
\end{figure*}

\begin{figure*}[tb]
    \begin{subfigure}[b]{0.99\linewidth}
        \centering
        \includegraphics[keepaspectratio=true,width=\linewidth]{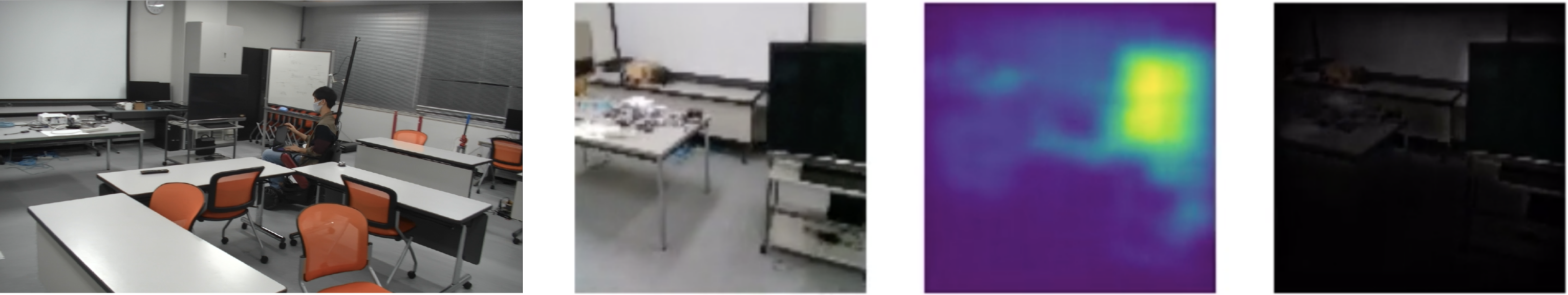}
        \subcaption{At corner}
        \label{fig:snap_tsallis_1}
    \end{subfigure}
    \begin{subfigure}[b]{0.99\linewidth}
        \centering
        \includegraphics[keepaspectratio=true,width=\linewidth]{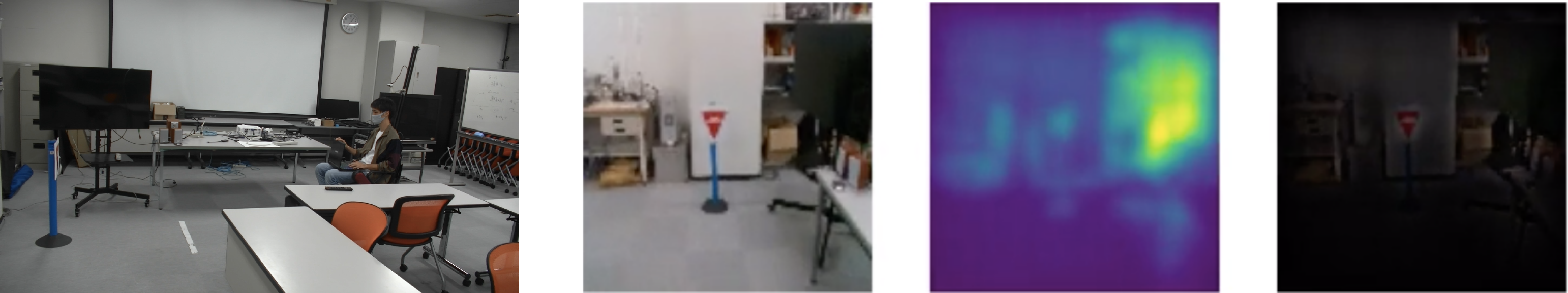}
        \subcaption{In front of stop sign}
        \label{fig:snap_tsallis_2}
    \end{subfigure}
    \begin{subfigure}[b]{0.99\linewidth}
        \centering
        \includegraphics[keepaspectratio=true,width=\linewidth]{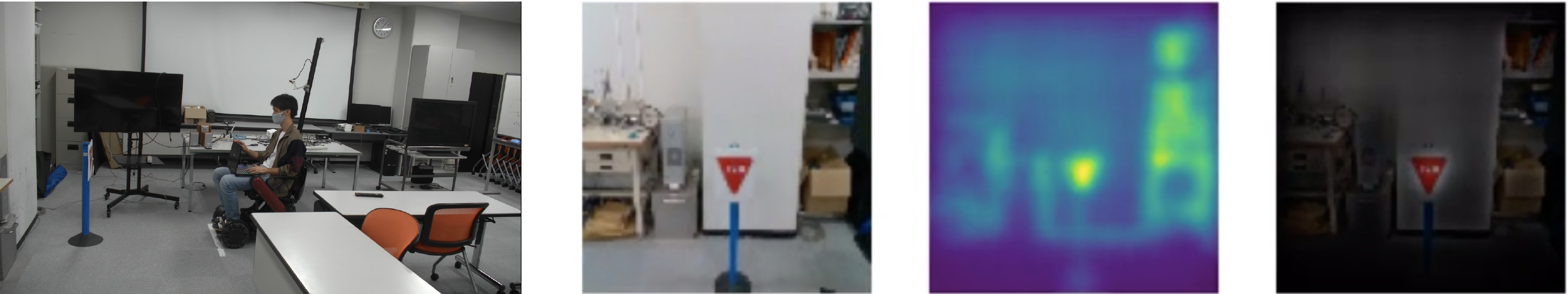}
        \subcaption{When stopped}
        \label{fig:snap_tsallis_3}
    \end{subfigure}
    \caption{Snapshots and the corresponding regions of interest when demonstrated by the proposed method:
    the wheelchair succeeded in turning the left at the corner smoothly, and it found the stop sign from a distance and successfully stopped just on the stop line.
    }
    \label{fig:snap_tsallis}
\end{figure*}

\begin{figure*}[tb]
    \begin{subfigure}[b]{0.245\linewidth}
        \centering
        \includegraphics[keepaspectratio=true,width=\linewidth]{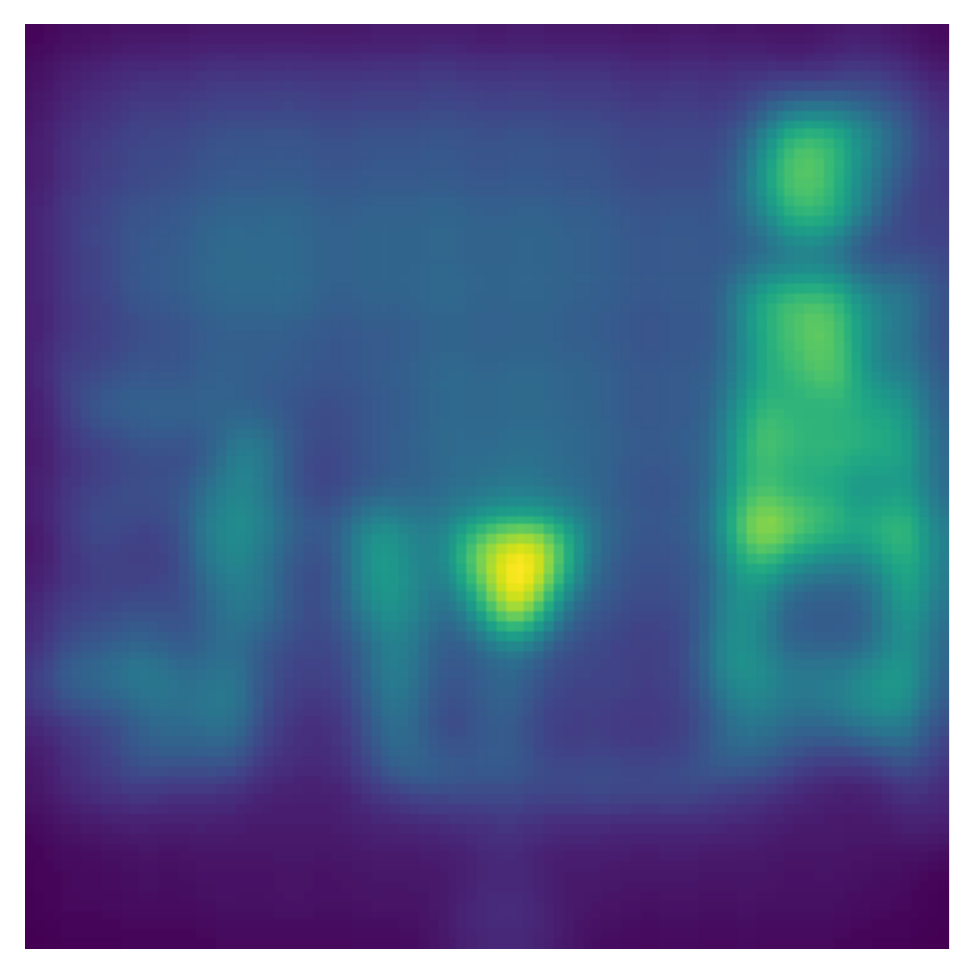}
        \subcaption{Original}
        \label{fig:vbp_base}
    \end{subfigure}
    \begin{subfigure}[b]{0.245\linewidth}
        \centering
        \includegraphics[keepaspectratio=true,width=\linewidth]{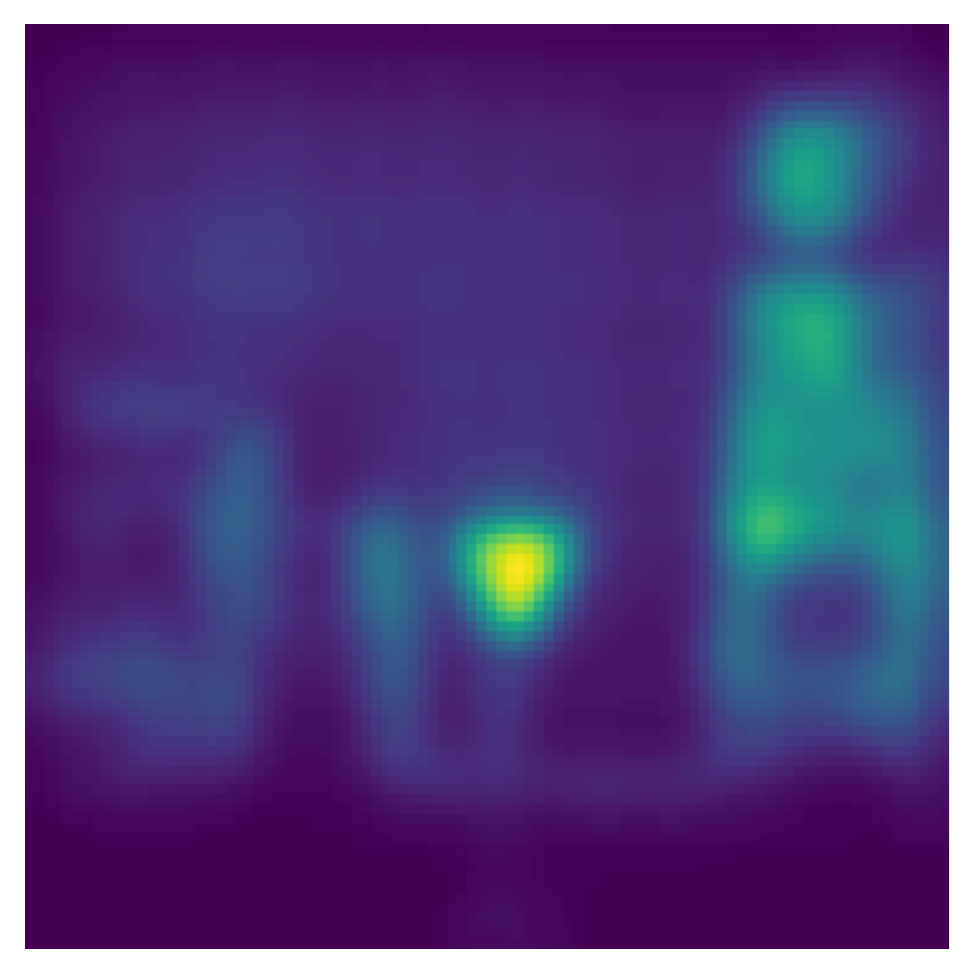}
        \subcaption{With normalization}
        \label{fig:vbp_norm}
    \end{subfigure}
    \begin{subfigure}[b]{0.245\linewidth}
        \centering
        \includegraphics[keepaspectratio=true,width=\linewidth]{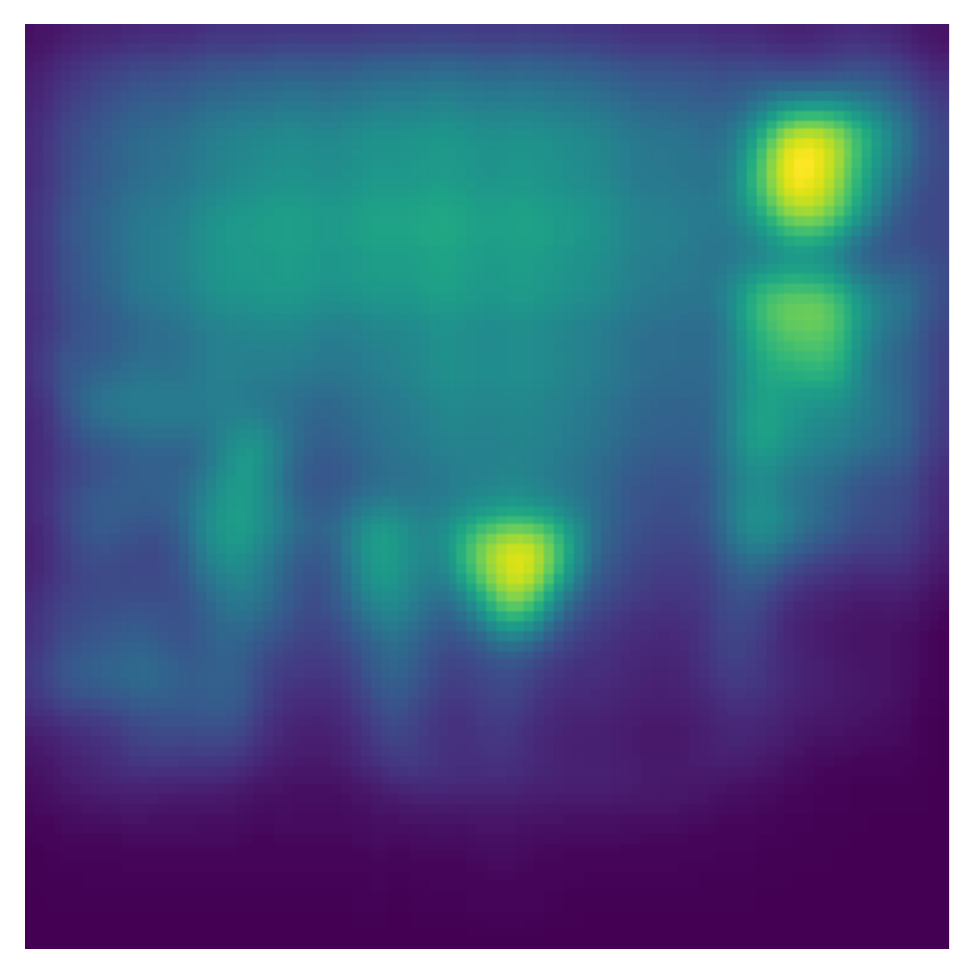}
        \subcaption{With consideration of FCNs}
        \label{fig:vbp_fc}
    \end{subfigure}
    \begin{subfigure}[b]{0.245\linewidth}
        \centering
        \includegraphics[keepaspectratio=true,width=\linewidth]{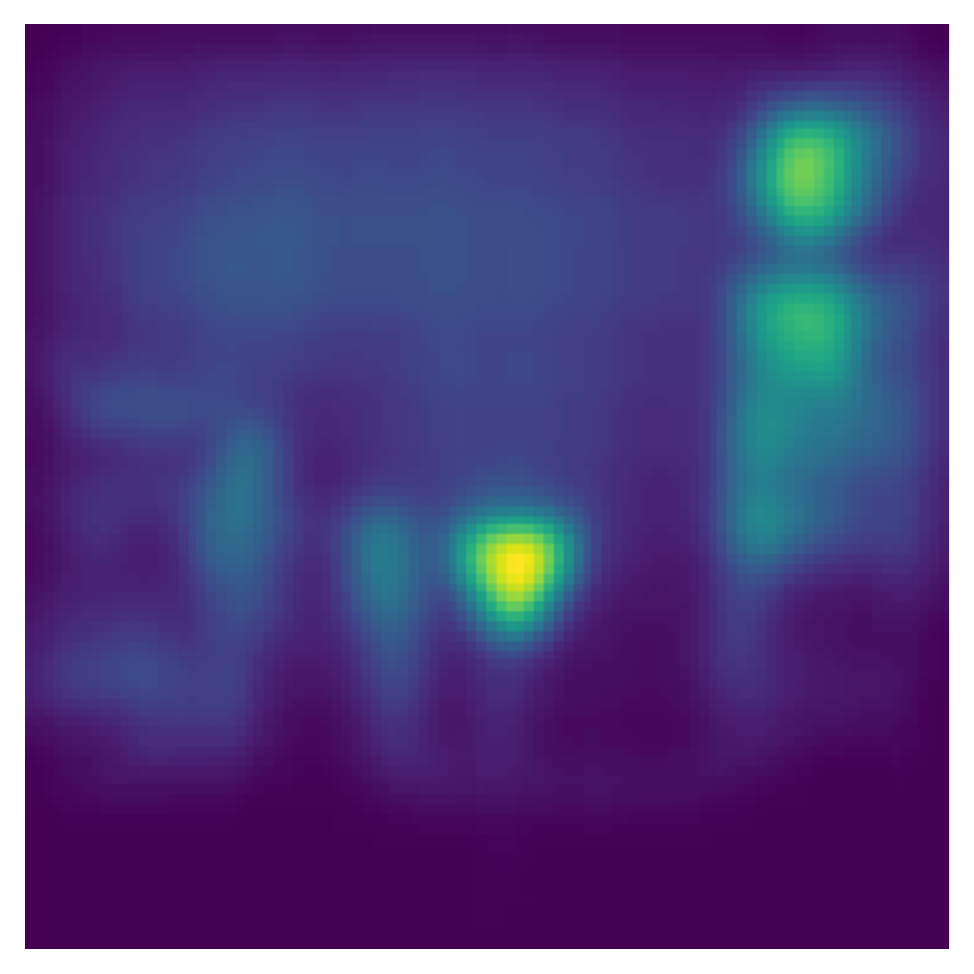}
        \subcaption{With both}
        \label{fig:vbp_both}
    \end{subfigure}
    \caption{Regions of interest extracted by the original and modified VisualBackProps:
    (a) the original is blurred overall;
    (b) by adding the normalization, the region of interest is clearer and emphasizes the stop sign more;
    (c) by adding the consideration of FCNs, the attention to the right shelf concentrated on the upper right;
    (d) by adding both, the region of interest is focused on the stop sign, which brings about stopping behavior, and on the upper right, which brings about normal operation in this corner, and we can interpret that the stronger stop sign yields stopping behavior.
    }
    \label{fig:vbp}
\end{figure*}

\subsection{Experimental setup}

\subsubsection{Dataset}

The proposed method is validated through an autonomous driving task of an electric wheelchair.
Our wheelchair is based on Whill Model CR with two cameras (Intel Realsense D435i) mounted overhead, as shown in Fig.~\ref{fig:exp_wheel}.
This wheelchair can control its translational and turning speeds by tilting a joystick in the hand back and forth, left and right.
The observation state is the RGB image acquired by the front camera and compressed to 96$\times$96 pixels, and the action is the two-dimensional operated values of the joystick, which can be given from ROS2~\cite{maruyama2016exploring} without operating the joystick.
Note that although two cameras were mounted on both the front and back sides, only the front camera was used in the experiment for simplicity.

When collecting the dataset, one driver attempted to drive the wheelchair clockwise and counterclockwise around a rectangular course.
A stop sign was placed at the second corner from the starting point, and one trajectory was defined until stopping in front of it for three seconds (see Fig.~\ref{fig:exp_course}).
State and action were stored at 50 fps, and in total, we collected 90 trajectories with eight patterns.
They were divided 82 trajectories with 21137 state-action pairs into the training dataset and the remaining eight trajectories with 1149 state-action pairs into the test dataset.
Here, to show robustness to noisy data, we intentionally mixed in two trajectories of the training data that did zigzag and/or not stop before the stop sign.

\subsubsection{Architecture}

For approximating the stochastic controller $\pi$, we combine CNNs and FCNs as described in Fig.~\ref{fig:arch_network}.
This architecture is implemented by PyTorch~\cite{paszke2017automatic}.
The activation function for each layer is the ReLU function, and we introduce InstanceNorm~\cite{ulyanov2016instance} for CNNs and LayerNorm~\cite{ba2016layer} for FCNs to stabilize learning.
Since the controller is modeled as a multivariate diagonal normal distribution, the more specific outputs from the architecture are the mean and variance parameters.

To train this architecture, we use Adam~\cite{kingma2014adam}, which is the most popular stochastic gradient descent optimizer, with a batch size of 512 and a learning rate of $10^{-5}$.
One epoch of training is to use all the training data once randomly, and the training is forcibly terminated after 100 epochs.
In order to take into account the randomness of the initialization, the training is performed three times for each condition and the mean of the results is used for comparison.

\subsection{Scores for test dataset}

First, we investigate the effect of $q$, a hyperparameter added in the proposed method.
The scores of training with $q$ with 0.1 decrements from $1$ are shown in Fig.~\ref{fig:result_qvals}.
Note that since the loss function is modified in the proposed method, the original negative log likelihood for the test dataset was employed as score.
As expected, we found that too small $q$ resulted in extremely poor score, since it excludes most of the data as noise and fits the remaining few.
On the other hand, for $q \in [0.7, 0.9]$, their scores were roughly the same as that of the conventional method ($q=1$), with a minimum at $q=0.8$.
In fact, comparing the learning curves with $q=0.8$ and $q=1$, Fig.~\ref{fig:result_loss} can see that the learning curve of the proposed method was noticeably lower than that of the conventional method.

From these results, we conclude that by specifying the appropriate $q$, the proposed method can increase the likelihood of the controller for the test dataset than the conventional method.
This fact indicates that while the conventional method updates the controller in the wrong direction to represent even the noisy data contained in the training dataset, the proposed method can properly exclude them and preferentially fit the data similar to the test dataset.
As a remark $q$ needs to be adjusted according to the problem, but the best result can be obtained without a large burden by various efficient meta-optimization methods~\cite{srinivas2010gaussian,salinas2020quantile,aotani2021meta} or even by the grid search as in this paper.

\subsection{Demonstrations}

We show examples of autonomous driving in which the controller learned by the conventional method (i.e. with $q=1$) or the proposed method at the best (i.e. with $q=0.8$) is deployed.
Details of the demonstrations can be found in the attached video.
Note that the region of interest here was visualized using the conventional VisualBackProp, where blue/yellow regions have the low/high attentions.

First, the demonstration using the conventional method is shown in Fig.~\ref{fig:snap_vanilla}.
It is easy to see in the video, but the joystick command moved noisily to the left and right even when going straight due to the effects of zigzagging.
In addition, the wheelchair did not pay attention to the stop sign when finding it, and failed to stop at an appropriate distance, which can be confirmed by the white line on the floor.
When the wheelchair stopped, it still paid attention to the shelf on the right (probably as a guide for going straight and/or turning left) instead of paying attention to the stop sign.
Therefore we can say that the imitation failed and the region of interest was clearly wrong, indicating that the influence of noisy data was strong.

In contrast, the demonstration using the proposed method was successfully completed, as shown in Fig.~\ref{fig:snap_tsallis}.
The joystick command was hardly noisy even when going straight ahead.
In addition, it can be seen that the wheelchair started to pay attention to the stop sign when finding it.
When the wheelchair finally stopped at the appropriate distance, it paid the most attention to the stop sign, indicating that it used this as a landmark for its stopping motion.
Thus, we can conclude that the proposed method did not get confused by noisy data, but relied on the optimal other data for successful imitation.

\subsection{Analysis by modified VisualBackProp}

As can be seen in Fig.~\ref{fig:snap_tsallis}, although VisualBackProp extracted the region of interest that seems to be natural, the whole image was blurred.
Therefore, we judged that its visualization would have room for improvements.
We examine the effects of the two proposed modifications, i.e. normalization and consideration of FCNs, on Fig.~\ref{fig:snap_tsallis_3}.

Fig.~\ref{fig:vbp} shows the regions of interest by the respective modifications.
First, it is noticeable that the region of interest was clearer with the normalization.
This allows us to judge with more confidence that the wheelchair was stopped by checking the stop sign.
Although the consideration of FCNs made the region of interest slightly more blurred (probably due to the insufficient sparseness), it can be seen that the emphasis on the entire right shelf was reduced to focus only on the upper right.
In fact, objects with unique colors in the course are placed in the upper right, suggesting that they can be easily used as landmarks for going straight and/or turning left.
By integrating these modifications, the region of interest could be limited to the stop sign and the objects in the upper right, while paying stronger attention to the stop sign.
This implies that the driver and the learned controller make a decision between stopping and going straight and/or turning left (in this corner) based on these two characteristic regions of interest, and that they shift to the stopping behavior when the stop sign is close enough.

As a consequence, the region of interest was clearer than that of the conventional VisualBackProp, and its contents were natural enough to be interpreted.
Therefore, it is suggested that the proposed modifications can surely improve the visualization performance.

\section{Discussion}
\label{sec:discussion}

\subsection{Limitations of the proposed method}

We experimentally confirmed that the proposed method can indeed achieve robust imitation against noise.
However, unless $q$ is appropriately tuned, the proposed method may collapse the controller, although meta-optimization is possible at relatively low cost, as mentioned before~\cite{srinivas2010gaussian,salinas2020quantile,aotani2021meta}.
In addition, it is not obvious whether there exists $q < 1$ that necessarily outperforms the conventional method.
Especially when the variance of the expert controller is large, or when the action space is discrete and there are many choices, $\pi\ll 1$ is basically satisfied, and almost all data can be with weights of less than 1 (ultimately 0).
In that cases, $q \simeq 1$ should be better to have the non-zero weights, but that would revert noise sensitivity again.
Therefore, although the proposed method is effective for autonomous driving tasks in which the control command is continuous and relatively deterministic, we have to carefully use the proposed method for imitating more general tasks.

Since the proposed method is formulated based on the standard behavioral cloning, it inherits the problems of behavioral cloning (except for the noise sensitivity).
For example, the open issues about covariate shift and compounding error are often discussed in the literature of imitation learning~\cite{laskey2017dart,brantley2020disagreement,ho2016generative}.
In the near future, the proposed method should be properly integrated with methods that mitigate these problems.

\subsection{Alternative interpretations of behavioral cloning}

Behavioral cloning is formulated by eq.~\eqref{eq:loss_vanilla}, but new optimization problems have been reported by reinterpreting it as another optimization problem~\cite{sasaki2021behavioral,ghasemipour2020divergence}.
Specifically, eq.~\eqref{eq:loss_vanilla} is equivalent to minimizing the following Kullback-Leibler divergence.
\begin{align}
    & \mathbb{E}_{s_n \sim p_e} \left[ \mathrm{KL}(\pi_{\mathrm{exp}}(a \mid s_n) || \pi(a \mid s_n; \theta)) \right]
    \nonumber \\
    =& \mathbb{E}_{s_n \sim p_e} \left[ - \int_a \pi_{\mathrm{exp}}(a \mid s_n) \ln \cfrac{\pi(a \mid s_n; \theta)}{\pi_{\mathrm{exp}}(a \mid s_n)} da \right]
    \nonumber \\
    =& - \mathbb{E}_{s_n \sim p_e, a_n \sim \pi_{\mathrm{exp}}} \left[ \ln \pi(a_n \mid s_n; \theta) \right]
    - \mathbb{E}_{s_n \sim p_e} \left[ H(\pi_{\mathrm{exp}}) \right]
    \nonumber \\
    \propto& - \mathbb{E}_{(s_n, a_n) \sim D} \left[ \ln \pi(a_n \mid s_n; \theta) \right]
    \label{eq:loss_kl}
\end{align}
where $\pi_{\mathrm{exp}}$ denotes the expert controller and $p_e$ is the stochastic dynamics of the environment.
The expectation operation for these two distributions can be replaced by that for the dataset $D$ by Monte Carlo approximation, and the entropy $H(\pi_{\mathrm{exp}})$ can be excluded due to irrelevance to the optimization problem.
As a result, the optimization problem for the standard behavioral cloning is obtained.

The proposed method with eq.~\eqref{eq:loss_robust} can be reinterpreted in the same way.
In Tsallis statistics, the $q$-deformed Kullback-Leibler divergence (or Tsallis divergence) is also defined with a similar but different form~\cite{nielsen2011closed,gil2013renyi}.
\begin{align}
    \mathrm{KL}_q(p_1(x) || p_2(x)) = - \int_x p_1(x) \ln_q \cfrac{p_2(x)}{p_1(x)} dx
    \label{eq:def_klq}
\end{align}
In addition, the decomposition of $q$-logarithm is specially given as follows:
\begin{align}
    \ln_q \cfrac{y}{x} = x^{q-1}(\ln_q y - \ln_q x)
    \label{eq:def_lnq_div}
\end{align}

With these two definitions, we derive the following minimization problem.
\begin{align}
    & \mathbb{E}_{s_n \sim p_e} \left[ \mathrm{KL}_q(\pi_{\mathrm{exp}}(a \mid s_n) || \pi(a \mid s_n; \theta)) \right]
    \nonumber \\
    =& \mathbb{E}_{s_n \sim p_e} \left[ - \int_a \pi_{\mathrm{exp}}(a \mid s_n) \ln_q \cfrac{\pi(a \mid s_n; \theta)}{\pi_{\mathrm{exp}}(a \mid s_n)} da \right]
    \nonumber \\
    =& - \mathbb{E}_{s_n \sim p_e, a_n \sim \pi_{\mathrm{exp}}} \left[ \pi_{\mathrm{exp}}^{q-1}(a_n \mid s_n) \ln_q \pi(a_n \mid s_n; \theta) \right]
    \nonumber \\
    -& \mathbb{E}_{s_n \sim p_e} \left[ H_q(\pi_{\mathrm{exp}}) \right]
    \nonumber \\
    \propto& - \mathbb{E}_{(s_n, a_n) \sim D} \left[ \pi_{\mathrm{exp}}^{q-1}(a_n \mid s_n) \ln_q \pi(a_n \mid s_n; \theta) \right]
    \label{eq:loss_klq}
\end{align}
where $H_q(\cdot)$ denotes Tsallis entropy, which can be excluded.
This is consistent with eq.~\eqref{eq:loss_robust}, except that it is multiplied by $\pi_{\mathrm{exp}}^{q-1}$.
However, since $\pi_{\mathrm{exp}}$ is unknown with some exceptions (see later), it must be removed somehow.

As the first removal method, we assume that $\pi_{\mathrm{exp}} \simeq \pi$.
With notice that we do not calculate the gradient of $\pi$ for this substitution, we have $\pi^{q-1} = \exp\{-(1 - q)\ln\pi\}$, which cancels out the gradient ratio $\rho$ that arises when considering the gradient of $\ln_q \pi$, as defined in eq.~\eqref{eq:ratio}.
Hence, under this assumption, the above optimization problem is perfectly consistent with the standard behavioral cloning.

As the second way, we assume that $\pi_{\mathrm{exp}} \simeq 1$, where the expert took all the actions with a constant likelihood to collect the dataset.
In this case, the above optimization problem is consistent with eq.~\eqref{eq:loss_robust}.
Hence, we can conclude that the proposed method is equivalent to the minimization problem of Tsallis divergence under the assumption of $\pi_{\mathrm{exp}} \simeq 1$.

This interpretation can be exploited, for example, to utilize Renyi divergence~\cite{nielsen2011closed,gil2013renyi} as a new minimization problem.
It can be transformed invertibly to Tsallis divergence, and it is possible that the gradient generated by the invertible transformation may provide different learning properties from the proposed method.

\subsection{Other applications of the proposed method}

While this paper utilized the property for $q < 1$ for the noise robustness, other applications can be discussed.
For example, if the task to be imitated has multiple correct solutions, the dataset will contain a wide variety of trajectories, and imitating all of them will require a very high level of approximation capability to CNNs and FCNs (and a model for the stochastic controller).
In such a case, the proposed method limits the number of trajectories to be imitated by excluding some of the various trajectories as noise, and thus it can be trained by a standard implementation.
This can be interpreted as the dataset distillation~\cite{wang2018dataset} in the loss function stage implicitly.

According to this interpretation, the proposed method should be effective in distilling the model~\cite{rusu2015policy,gou2021knowledge}.
Ideally, the distilled model should have the same level of performance as the original, but depending on its size, some performance degradation is inevitable.
In such a case, the proposed method would be able to achieve a distillation that excludes selectively some of the features but retains the rest, rather than degrading the overall performance.

As a remark, in the above minimization problem of Tsallis divergence, $\pi_{\mathrm{exp}}$ was assumed to be unknown in general, but $\pi_{\mathrm{exp}}$ can be revealed in the model distillation.
In this case, for the standard behavioral cloning, the gradient ratio is given by $\exp\{(1 - q)(\ln\pi - \ln\pi_{\mathrm{exp}})\}$.
That is, the weighting is relative in this form, whereas it was absolute in the proposed method.
Although the expected behavior is similar, it will be possible to prioritize relatively important data by appropriately weighting cases where the variance of $\pi_{\mathrm{exp}}$ is large and the entire data tends to be ignored.

\section{Conclusion}
\label{sec:conclusion}

In this paper, we proposed a novel behavioral cloning method based on Tsallis statistics that is robust to the small and noisy personal dataset especially in the automated personal mobility task.
Specifically, we focused on that the standard behavioral cloning utilizes the log likelihood of the stochastic controller, and replaced it with the $q$-log likelihood.
We showed analytically that this replacement provides the noise robustness.
We also identified minor issues with VisualBackProp, which is useful for visually verifying task performance, and implemented the ad-hoc solutions, i.e. the normalization of all the features and the consideration of FCNs.
With the experimental results, it can be concluded that the proposed method can learn correctly even by the dataset that conventionally fail to be imitated, and has the similar region of interest to the driver.

In the future, we aim to conduct larger-scale experiments and further improve imitation learning based on Tsallis statistics.
In particular, we would like to investigate and analyze whether this concept can be successfully used to solve covariate shift and compounding error, which are open issues in behavioral cloning.

\section*{Acknowledgements}

This work was supported by The Support Center for Advanced Telecommunications Technology Research Foundation (SCAT) Research Grant.

%
%
%
\bibliographystyle{elsarticle-num}
\bibliography{biblio}	

\begin{thebibliography}{10}
\expandafter\ifx\csname url\endcsname\relax
  \def\url#1{\texttt{#1}}\fi
\expandafter\ifx\csname urlprefix\endcsname\relax\def\urlprefix{URL }\fi
\expandafter\ifx\csname href\endcsname\relax
  \def\href#1#2{#2} \def\path#1{#1}\fi

\bibitem{suryanarayanan2007appropriate}
S.~Suryanarayanan, M.~Tomizuka, Appropriate sensor placement for fault-tolerant
  lane-keeping control of automated vehicles, IEEE/ASME Transactions on
  mechatronics 12~(4) (2007) 465--471.

\bibitem{klanvcar2009wheeled}
G.~Klan{\v{c}}ar, D.~Matko, S.~Bla{\v{z}}i{\v{c}}, Wheeled mobile robots
  control in a linear platoon, Journal of Intelligent and Robotic Systems
  54~(5) (2009) 709--731.

\bibitem{levinson2011towards}
J.~Levinson, J.~Askeland, J.~Becker, J.~Dolson, D.~Held, S.~Kammel, J.~Z.
  Kolter, D.~Langer, O.~Pink, V.~Pratt, et~al., Towards fully autonomous
  driving: Systems and algorithms, in: 2011 IEEE intelligent vehicles symposium
  (IV), IEEE, 2011, pp. 163--168.

\bibitem{williams2018information}
G.~Williams, P.~Drews, B.~Goldfain, J.~M. Rehg, E.~A. Theodorou,
  Information-theoretic model predictive control: Theory and applications to
  autonomous driving, IEEE Transactions on Robotics 34~(6) (2018) 1603--1622.

\bibitem{codevilla2018end}
F.~Codevilla, M.~M{\"u}ller, A.~L{\'o}pez, V.~Koltun, A.~Dosovitskiy,
  End-to-end driving via conditional imitation learning, in: IEEE International
  Conference on Robotics and Automation, IEEE, 2018, pp. 4693--4700.

\bibitem{onishi2019end}
T.~Onishi, T.~Motoyoshi, Y.~Suga, H.~Mori, T.~Ogata, End-to-end learning method
  for self-driving cars with trajectory recovery using a path-following
  function, in: International Joint Conference on Neural Networks, IEEE, 2019,
  pp. 1--8.

\bibitem{hawke2020urban}
J.~Hawke, R.~Shen, C.~Gurau, S.~Sharma, D.~Reda, N.~Nikolov, P.~Mazur,
  S.~Micklethwaite, N.~Griffiths, A.~Shah, et~al., Urban driving with
  conditional imitation learning, in: IEEE International Conference on Robotics
  and Automation, IEEE, 2020, pp. 251--257.

\bibitem{codevilla2019exploring}
F.~Codevilla, E.~Santana, A.~M. L{\'o}pez, A.~Gaidon, Exploring the limitations
  of behavior cloning for autonomous driving, in: IEEE/CVF International
  Conference on Computer Vision, 2019, pp. 9329--9338.

\bibitem{nakajima2017new}
S.~Nakajima, A new personal mobility vehicle for daily life: improvements on a
  new rt-mover that enable greater mobility are showcased at the cybathlon,
  IEEE Robotics \& Automation Magazine 24~(4) (2017) 37--48.

\bibitem{nguyen2004segway}
H.~G. Nguyen, J.~Morrell, K.~D. Mullens, A.~B. Burmeister, S.~Miles,
  N.~Farrington, K.~M. Thomas, D.~W. Gage, Segway robotic mobility platform,
  in: Mobile Robots XVII, Vol. 5609, International Society for Optics and
  Photonics, 2004, pp. 207--220.

\bibitem{argall2009survey}
B.~D. Argall, S.~Chernova, M.~Veloso, B.~Browning, A survey of robot learning
  from demonstration, Robotics and autonomous systems 57~(5) (2009) 469--483.

\bibitem{hussein2017imitation}
A.~Hussein, M.~M. Gaber, E.~Elyan, C.~Jayne, Imitation learning: A survey of
  learning methods, ACM Computing Surveys 50~(2) (2017) 1--35.

\bibitem{wu2019imitation}
Y.-H. Wu, N.~Charoenphakdee, H.~Bao, V.~Tangkaratt, M.~Sugiyama, Imitation
  learning from imperfect demonstration, in: International Conference on
  Machine Learning, PMLR, 2019, pp. 6818--6827.

\bibitem{tangkaratt2020variational}
V.~Tangkaratt, B.~Han, M.~E. Khan, M.~Sugiyama, Variational imitation learning
  with diverse-quality demonstrations, in: International Conference on Machine
  Learning, PMLR, 2020, pp. 9407--9417.

\bibitem{bain1995framework}
M.~Bain, C.~Sammut, A framework for behavioural cloning., in: Machine
  Intelligence 15, 1995, pp. 103--129.

\bibitem{ng2000algorithms}
A.~Y. Ng, S.~J. Russell, Algorithms for inverse reinforcement learning, in:
  International Conference on Machine Learning, 2000, pp. 663--670.

\bibitem{sutton2018reinforcement}
R.~S. Sutton, A.~G. Barto, Reinforcement learning: An introduction, MIT press,
  2018.

\bibitem{hussein2021robust}
M.~Hussein, B.~Crowe, M.~Clark-Turner, P.~Gesel, M.~Petrik, M.~Begum, Robust
  behavior cloning with adversarial demonstration detection, in: IEEE/RSJ
  International Conference on Intelligent Robots and Systems, IEEE, 2021, pp.
  7835--7841.

\bibitem{sasaki2021behavioral}
F.~Sasaki, R.~Yamashina, Behavioral cloning from noisy demonstrations, in:
  International Conference on Learning Representations, 2021.

\bibitem{ilboudo2020robust}
W.~E.~L. Ilboudo, T.~Kobayashi, K.~Sugimoto, Robust stochastic gradient descent
  with student-t distribution based first-order momentum, IEEE Transactions on
  Neural Networks and Learning Systems.

\bibitem{ilboudo2021adaptive}
W.~E.~L. Ilboudo, T.~Kobayashi, K.~Sugimoto, Adaptive t-momentum-based
  optimization for unknown ratio of outliers in amateur data in imitation
  learning, in: IEEE/RSJ International Conference on Intelligent Robots and
  Systems, IEEE, 2021, pp. 7828--7834.

\bibitem{tsallis1988possible}
C.~Tsallis, Possible generalization of boltzmann-gibbs statistics, Journal of
  statistical physics 52~(1-2) (1988) 479--487.

\bibitem{suyari2005law}
H.~Suyari, M.~Tsukada, Law of error in tsallis statistics, IEEE Transactions on
  Information Theory 51~(2) (2005) 753--757.

\bibitem{kobayashi2020q}
T.~Kobayashi, q-vae for disentangled representation learning and latent
  dynamical systems, IEEE Robotics and Automation Letters 5~(4) (2020)
  5669--5676.

\bibitem{bojarski2018visualbackprop}
M.~Bojarski, A.~Choromanska, K.~Choromanski, B.~Firner, L.~J. Ackel, U.~Muller,
  P.~Yeres, K.~Zieba, Visualbackprop: Efficient visualization of cnns for
  autonomous driving, in: IEEE International Conference on Robotics and
  Automation, IEEE, 2018, pp. 4701--4708.

\bibitem{krizhevsky2012imagenet}
A.~Krizhevsky, I.~Sutskever, G.~E. Hinton, Imagenet classification with deep
  convolutional neural networks, in: Advances in neural information processing
  systems, 2012, pp. 1097--1105.

\bibitem{lecun2015deep}
Y.~LeCun, Y.~Bengio, G.~Hinton, Deep learning, nature 521~(7553) (2015) 436.

\bibitem{kingma2014adam}
D.~P. Kingma, J.~Ba, Adam: A method for stochastic optimization, arXiv preprint
  arXiv:1412.6980.

\bibitem{zeiler2011adaptive}
M.~D. Zeiler, G.~W. Taylor, R.~Fergus, Adaptive deconvolutional networks for
  mid and high level feature learning, in: 2011 International Conference on
  Computer Vision, IEEE, 2011, pp. 2018--2025.

\bibitem{maruyama2016exploring}
Y.~Maruyama, S.~Kato, T.~Azumi, Exploring the performance of ros2, in:
  International Conference on Embedded Software, 2016, pp. 1--10.

\bibitem{paszke2017automatic}
A.~Paszke, S.~Gross, S.~Chintala, G.~Chanan, E.~Yang, Z.~DeVito, Z.~Lin,
  A.~Desmaison, L.~Antiga, A.~Lerer, Automatic differentiation in pytorch, in:
  Advances in Neural Information Processing Systems Workshop, 2017.

\bibitem{ulyanov2016instance}
D.~Ulyanov, A.~Vedaldi, V.~Lempitsky, Instance normalization: The missing
  ingredient for fast stylization, arXiv preprint arXiv:1607.08022.

\bibitem{ba2016layer}
J.~L. Ba, J.~R. Kiros, G.~E. Hinton, Layer normalization, arXiv preprint
  arXiv:1607.06450.

\bibitem{srinivas2010gaussian}
N.~Srinivas, A.~Krause, S.~Kakade, M.~Seeger, Gaussian process optimization in
  the bandit setting: no regret and experimental design, in: International
  Conference on International Conference on Machine Learning, Omnipress, 2010,
  pp. 1015--1022.

\bibitem{salinas2020quantile}
D.~Salinas, H.~Shen, V.~Perrone, A quantile-based approach for hyperparameter
  transfer learning, in: International Conference on Machine Learning, PMLR,
  2020, pp. 8438--8448.

\bibitem{aotani2021meta}
T.~Aotani, T.~Kobayashi, K.~Sugimoto, Meta-optimization of bias-variance
  trade-off in stochastic model learning, IEEE Access 9 (2021) 148783--148799.

\bibitem{laskey2017dart}
M.~Laskey, J.~Lee, R.~Fox, A.~Dragan, K.~Goldberg, Dart: Noise injection for
  robust imitation learning, in: Conference on robot learning, PMLR, 2017, pp.
  143--156.

\bibitem{brantley2020disagreement}
K.~Brantley, W.~Sun, M.~Henaff, Disagreement-regularized imitation learning,
  in: International Conference on Learning Representations, 2020.

\bibitem{ho2016generative}
J.~Ho, S.~Ermon, Generative adversarial imitation learning, Advances in neural
  information processing systems 29 (2016) 4565--4573.

\bibitem{ghasemipour2020divergence}
S.~K.~S. Ghasemipour, R.~Zemel, S.~Gu, A divergence minimization perspective on
  imitation learning methods, in: Conference on Robot Learning, PMLR, 2020, pp.
  1259--1277.

\bibitem{nielsen2011closed}
F.~Nielsen, R.~Nock, A closed-form expression for the sharma--mittal entropy of
  exponential families, Journal of Physics A: Mathematical and Theoretical
  45~(3) (2011) 032003.

\bibitem{gil2013renyi}
M.~Gil, F.~Alajaji, T.~Linder, R{\'e}nyi divergence measures for commonly used
  univariate continuous distributions, Information Sciences 249 (2013)
  124--131.

\bibitem{wang2018dataset}
T.~Wang, J.-Y. Zhu, A.~Torralba, A.~A. Efros, Dataset distillation, arXiv
  preprint arXiv:1811.10959.

\bibitem{rusu2015policy}
A.~A. Rusu, S.~G. Colmenarejo, C.~Gulcehre, G.~Desjardins, J.~Kirkpatrick,
  R.~Pascanu, V.~Mnih, K.~Kavukcuoglu, R.~Hadsell, Policy distillation, arXiv
  preprint arXiv:1511.06295.

\bibitem{gou2021knowledge}
J.~Gou, B.~Yu, S.~J. Maybank, D.~Tao, Knowledge distillation: A survey,
  International Journal of Computer Vision 129~(6) (2021) 1789--1819.

\end{thebibliography}



\end{document}